\documentclass[11pt]{article}
\pdfoutput=1

\usepackage[preprint]{coling}

\usepackage{times}
\usepackage{latexsym}

\usepackage[T1]{fontenc}

\usepackage[utf8]{inputenc}

\usepackage{microtype}

\usepackage{inconsolata}

\usepackage{graphicx}

\usepackage{multirow}
\usepackage{xspace}
\usepackage{booktabs}
\usepackage{todonotes}

\usepackage[inline]{enumitem}

\usepackage{amsmath}
\DeclareMathOperator*{\argmax}{arg\,max}

\usepackage[acronym]{glossaries}
\makenoidxglossaries 
\usepackage{float}
\usepackage{subcaption}

\usepackage{fontawesome}
\usepackage{supertabular}
\usepackage{siunitx}[=v2]

\newacronym{NLP}{NLP}{Natural Language Processing}
\newacronym{NMT}{NMT}{Neural Machine Translation}
\newacronym{MT}{MT}{Machine Translation}

\newacronym{LM}{LM}{Language Model}
\newacronym{LLM}{LLM}{Large Language Model}
\newacronym{MLLM}{MLLM}{Multilingual Large Language Model}
\newacronym{MoE}{MoE}{Mixture-of-Experts}
\newacronym{SMoE}{SMoE}{Sparse Mixture-of-Experts}

\newacronym{ML}{ML}{Machine Learning}
\newacronym{FFN}{FFN}{Feed Forward Network}

\newacronym{NEL}{NEL}{Named Entity Linking}
\newacronym{NED}{NED}{Named Entity Disambiguation}
\newacronym{NER}{NER}{Named Entity Recognition}
\newacronym{NE}{NE}{Named Entity}
\newacronym{NEs}{NEs}{Named Entities}

\newacronym{mKGQA}{mKGQA}{Multilingual Knowledge Graph Question Answering}
\newacronym{KGQA}{KGQA}{Knowledge Graph Question Answering}
\newacronym{QALDPlus}{QALD-9-plus}{QALD-9-Plus Dataset}
\newacronym{QALD}{QALD}{Question Answering over Linked Data}
\newacronym{QA}{QA}{Question Answering}
\newacronym{KG}{KG}{Knowledge Graph}
\newacronym{KGs}{KGs}{Knowledge Graphs}
\newacronym{RDF}{RDF}{Resource Description Framework}
\newacronym{NEAMT}{NEAMT}{Named Entity Aware Machine Translation}
\newacronym{SPARQL}{SPARQL}{SPARQL}
\newacronym{QAnswer}{QAnswer}{QAnswer}
\newacronym{URI}{URI}{Uniform Resource Identifier}
\newacronym{IR}{IR}{Information Retrieval}
\newacronym{NsPM}{NsPM}{Neural SPARQL Machines}

\newacronym{FOSS}{FOSS}{free and open-source software}
\newacronym{FLOSS}{FLOSS}{free/libre and open-source software}

\newacronym{API}{API}{Application Programming Interface}

\newcommand{\ie}{i.e.,~}

\newcommand{\eg}{e.g.,~}

\newcommand{\llms}{{\glspl{LLM}}\xspace} 
\newcommand{\moe}{{\acrshort{MoE}}\xspace} 
\newcommand{\moefull}{{\acrfull{MoE}}\xspace} 
\newcommand{\com}{\textit{the curse of multilinguality}\xspace} 
\newcommand{\ffns}{{\glspl{FFN}}\xspace} 

\newcommand{\link}{\small{\faExternalLink{}}}

\newcommand{\approach}{LOLA\xspace}

\title{LOLA -- An Open-Source Massively Multilingual Large Language Model}

\author{
\bf Nikit Srivastava$^1$, Denis Kuchelev, Tatiana Moteu Ngoli$^1$, Kshitij Shetty$^2$,
\\ \bf Michael Röder$^1$, Hamada M. Zahera$^1$, Diego Moussallem$^1$, Axel-Cyrille Ngonga Ngomo$^1$  
\\ \\ Data Science Group, Paderborn University, Germany  
\\ \\ $^1$ \{nikit.srivastava, tatiana.moteu, michael.roeder, \\hamada.zahera, diego.moussallem, axel.ngonga\}@upb.de \\
$^2$ kshitij@mail.upb.de \\
}

\begin{document}
\maketitle
\begin{abstract}

This paper presents LOLA, a massively multilingual large language model trained on more than 160 languages using a sparse Mixture-of-Experts Transformer architecture. Our architectural and implementation choices address the challenge of harnessing linguistic diversity while maintaining efficiency and avoiding the common pitfalls of multilinguality. Our analysis of the evaluation results shows competitive performance in natural language generation and understanding tasks. Additionally, we demonstrate how the learned expert-routing mechanism exploits implicit phylogenetic linguistic patterns to potentially alleviate the curse of multilinguality. We provide an in-depth look at the training process, an analysis of the datasets, and a balanced exploration of the model’s strengths and limitations. As an open-source model, LOLA promotes reproducibility and serves as a robust foundation for future research. Our findings enable the development of compute-efficient multilingual models with strong, scalable performance across languages.

\end{abstract}

\section{Introduction}

\llms have shown tremendous capability across a diverse set of tasks in recent years~\cite{radford2019language, kaddour2023challengesapplicationslargelanguage}. This progress has propelled research, with many chat-based \llms
\footnote{\begin{minipage}[t]{\linewidth}
ChatGPT: \href{https://chat.openai.com/}{chat.openai.com}; \\
LLAMA: \href{https://llama.meta.com/}{llama.meta.com}; \\
Mistral: \href{https://mistral.ai/}{mistral.ai}; \\
Gemini: \href{https://gemini.google.com/}{gemini.google.com}; \\
Deepseek: \href{https://www.deepseek.com/}{deepseek.com}.
\end{minipage}}
gaining popularity among general users. However, concerns remain, particularly regarding their accessibility for  multilingual usage~\cite{joshi-etal-2020-state} and open-source licensing policies~\cite{liu2023llm360fullytransparentopensource}.
The number of competent \llms significantly decreases for languages other than English~\cite{2024ayamodelinstructionfinetuned}.
This, combined with \com---a phenomenon in which the ability of models to generalize across multiple languages diminishes unless their capacity is significantly expanded~\cite{conneau-etal-2020-unsupervised}---means non-English speakers often have access to inferior systems.
Additionally, many new models~\cite{jiang2023mistral7b, openai2024gpt4technicalreport, dubey2024llama3herdmodels} are pay-to-use, require personal information, or do not fully disclose training details, creating significant hurdles for multilingual research.

To advance multilingual language modeling, we introduce \approach,\footnote{
\begin{minipage}[t]{\linewidth}
Source Code: \href{https://github.com/dice-group/LOLA}{github.com/dice-group/LOLA}; \\
Model Weights: \href{https://huggingface.co/dice-research/lola_v1}{huggingface.co/dice-research/lola\_v1}.
\end{minipage}
} a massively multilingual model that follows a GPT-style~\cite{radford2019language} decoder-only architecture with sparse \acrfull{MoE} layers~\cite{shazeer2017outrageouslylargeneuralnetworks}. \moe architectures have shown strong performance on learning the underlying structure of the data ~\cite{NEURIPS2022_91edff07} but their application in multilingual \llms remains underexplored. 
\moe models can effectively increase model capacity with minimal additional computational cost, offering the possibility of leveraging implicit clusters like language family groups and playing a crucial role in addressing the challenges of multilinguality.

Language family groups, consisting of languages sharing common ancestral roots, offer opportunities for enhancing language models. Despite linguistic diversity, these families exhibit structural, syntactic, and semantic similarities~\cite{rowe2015concise} that can be exploited to improve performance across related languages.
Our goal is to leverage \moe's strengths to exploit the phylogenetic structure of languages and achieve better prediction performance. In particular, the shared and non-shared parameters of \moe-based models offer a promising approach to mitigating \com by increasing capacity while remaining compute efficient~\cite{shazeer2017outrageouslylargeneuralnetworks}. By exploiting language families, we aim to close gaps in current models---particularly for low-resource languages---by enhancing cross-linguistic transfer learning.

Another important factor is the availability of \llms as a free resource, accessible for anyone to use, modify, and redistribute without discrimination against any individuals or purposes. 
Many popular \llms that claim to be "open source" either withhold their training datasets (\eg Mistral, Grok\footnote{\href{https://github.com/xai-org/grok-1}{github.com/xai-org/grok-1}}), fail to publish their training code (\eg Llama, Grok), or do not release their inference code (\eg Grok-2\footnote{\href{https://x.ai/blog/grok-2}{x.ai/blog/grok-2}})~\cite{spectrum2024llm}. In some cases, these models are released under licenses that are restrictive, discriminatory, or impose additional conditions \cite{10.1145/3571884.3604316, 10.1145/3630106.3659005}. The artifacts and components used in \approach were selected based on their suitability for 
training massively multilingual \llms while minimizing licensing concerns. All chosen components are obtainable, modifiable, and redistributable in accordance with the terms of their original licenses.

To assess \approach's performance, we evaluated it on four 
task types: \begin{enumerate*} \item Question Answering (Q\&A), \item Reasoning, \item Natural Language Inference (NLI), and \item Reading Comprehension. \end{enumerate*} In total, we assessed the model across 13 multilingual tasks, comparing it to 17 other models grouped into three categories based on their active parameter count.\footnote{The number of parameters a model utilizes per token \cite{fedus2022switchtransformersscalingtrillion}. This distinction is crucial for understanding the efficiency and performance of MoE models.} Our results demonstrate strong performance across most tasks, though we note the limitations in \begin{enumerate*} \item tasks involving factual and mathematical Q\&A; and \item comparisons with models that use more than five times the active parameters 
of \approach. \end{enumerate*} These findings are discussed in detail later in the paper.

Beyond presenting the multilingual model as our main contribution, we 
address the following key research questions: 
\begin{enumerate} 
\item \textit{Does training a Mixture-of-Experts model on a wide variety of languages enhance generalization or lead to confusion?}\item \textit{How do experts impact the model's capacity to leverage implicit language groups?} 
\item \textit{What are the potential limitations?} 
\end{enumerate}

\section{Related Work}

The development of \glspl{LLM} has gained significant momentum since the introduction of the Transformer architecture by \citet{vaswani2017}. As \llms grew in size and complexity, their capacity to model increasingly nuanced linguistic patterns expanded. Models like GPT3 and Llama \cite{ brown2020languagemodelsfewshotlearners, touvron2023llamaopenefficientfoundation} showcased the ability of large models to perform few-shot learning, a significant milestone that further highlighted the flexibility of Transformer-based architectures. As the need to extend their capabilities to handle multiple languages effectively became increasingly apparent, research into multilingual \llms surged, aiming to enable performance across diverse languages with a single model, reducing the need for language-specific systems~\cite{zhu-etal-2024-multilingual}. Key efforts in this area include systems such as mBERT, XLM-R, mT5, and BLOOM \cite{devlin2019bert, conneau-etal-2020-unsupervised, xue2021mt5, bloom2022}, with more recent models like Tower, SeaLLM, and Breeze \cite{alves2024tower, nguyen-etal-2024-seallms, hsu2024breeze7btechnicalreport} focusing on adapting primarily English-pretrained models into multilingual ones through continued training. However, research in multilingual \llms faces several challenges, particularly in balancing performance across languages while keeping training costs manageable, as emphasized by \citet{conneau-etal-2020-unsupervised}.

One of the significant challenges with scaling \llms is the computational cost associated with training and deploying models with billions or trillions of parameters. To address this, the \acrfull{MoE} paradigm has emerged as a promising approach for efficiently scaling large models. The \moe architecture proposed by \citet{shazeer2017outrageouslylargeneuralnetworks} introduces the concept of sparsity, where only a subset of the model’s parameters is activated during each forward pass, thereby reducing the computational burden while maintaining high performance. Their approach demonstrated that models could achieve state-of-the-art performance while being computationally efficient. Later approaches, such as GShard and Switch Transformers \cite{lepikhin2021gshard, fedus2022switchtransformersscalingtrillion}, extended the \moe framework by simplifying routing and enhancing scalability, enabling models with over a trillion parameters while maintaining efficient computational costs and setting new benchmarks in large-scale model training. These advances led to increased research in \moe-based \llms, resulting in models like GLaM, DeepSpeed MoE and Mixtral \cite{pmlr-v162-du22c, deepspeedmoe2022, jiang2024mixtralexperts}.

Given the unique architecture of the \moe-based \llms, \acrfull{MT} models have explored its potential in language grouping. Several \acrshort{MT} systems, such as M2M, NLLB, and Lingual-SMoE \cite{m2m,nllbteam2022languageleftbehindscaling,zhao2024sparse}, have trained MoE-based models to enable many-to-many translation, leveraging either learned or custom expert-routing mechanisms that assigns experts based on the language. Systems like NLLB continue to demonstrate state-of-the-art \acrshort{MT} performance to this day \cite{zhu-etal-2024-multilingual}. In the case of pre-trained base models, \citet{zoph2022stmoe} briefly touch upon the multilingual nature of \moe models, though they primarily note that expert load balancing loss constrains the model's capacity to assign language-specific experts. Despite these advances, the application of \moe for pre-training massively multilingual \llms remains underexplored. This research contributes to addressing that gap.
\section{Model Overview}
\begin{figure*}[t]
\centering
    \includegraphics[width=\textwidth]{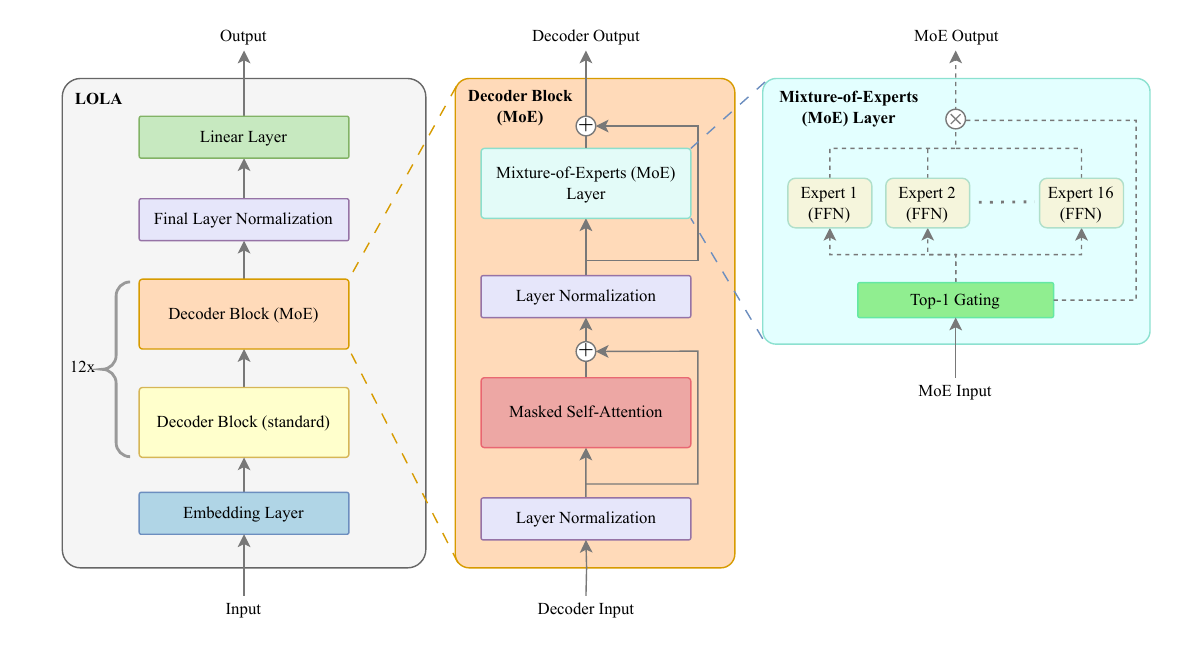}
    \caption{Three-level overview of the LOLA architecture. The left-most block provides a high-level overview of the layers within LOLA, including the alternating standard and \moefull-based decoder blocks. The middle block gives a detailed view of the \moe-based decoder block structure. The right-most block zooms in on the inner workings of each \moe layer, showing how the top-1 gating mechanism selects from multiple expert \ffns.    }
      \label{fig:lola-arch}
    \end{figure*}
Our model is based on a GPT-style \cite{radford2019language} decoder-only Transformer architecture \cite{vaswani2017}. We replace the standard feed-forward layers (\ffns) with \moefull layers in every alternate Transformer layer. These \moe layers utilize a \emph{top-1 gating} mechanism inspired by the Switch Transformer \cite{fedus2022switchtransformersscalingtrillion} due to its simplicity and effectiveness. The architecture consists of 24 decoder layers with a model hidden and embedding dimension of 2048, 16 attention heads, a maximum sequence length of 2048, and each \moe layer includes 16 experts. We use the GELU \cite{hendrycks2017bridging} non-linearities and the Adam \cite{adamopt} optimizer for our model. Based on this configuration, our model has 1.3 billion active parameters out of 7.4 billion total parameters. Due to this sparsity, our model has a training/inference cost similar to that of a 1.3 billion dense model.\footnote{Number of parameters activated in a single forward and backward pass.} \autoref{fig:lola-arch} provides a multi-level overview of the model architecture. The model configuration and training are facilitated using the Megatron-DeepSpeed\footnote{\href{https://github.com/microsoft/Megatron-DeepSpeed}{github.com/microsoft/Megatron-DeepSpeed}} framework, which is based on  \citet{shoeybi2020megatronlmtrainingmultibillionparameter, deepspeedmoe2022}.

\subsection{Routing Mechanism in MoE Layers}

For routing tokens through the \moe layers with \( N \) (\ie 16) experts, we first compute the logits for the gating function. These logits are then passed through a \textit{Softmax} function to calculate the probability for each expert:
\begin{equation}
    h(x) = W_g \cdot x,
    \label{eq:moe_logit}
\end{equation}
\begin{equation}
    G_i(x) = \frac{\exp(h(x)_i)}{\sum_{j=1}^{N} \exp(h(x)_j)},
    \label{eq:moe_gating}
\end{equation}
where \( h(x) \) contains the logit vectors for all experts, \( W_g \) is the gating weight matrix, and \( x \) is the input. The logit vector and gating probability of the \( i \)-th expert is denoted by \( h(x)_i \) and \( G_i(x) \) respectively.

Once the gating probabilities are computed, the output of the \moe layer is calculated by selecting the most probable expert \( i^* \) and multiplying its gating probability \( G_{i^*}(x) \) with the output of the corresponding expert \( E_{i^*}(x) \):
\begin{equation}
    i^* = \argmax_i G_i(x) \label{eq:argmax},
\end{equation}
\begin{equation}
    \text{MoE}(x) = G_{i^*}(x) \cdot E_{i^*}(x) \label{eq:moe_out}.
\end{equation}

\subsection{Training and Loss Functions}

Our model is pre-trained using a causal language modeling task \cite{Radford2018ImprovingLU}, where the objective is to minimize the cross-entropy loss alongside an auxiliary \moe loss.This auxiliary loss, inspired by works such as \citet{shazeer2017outrageouslylargeneuralnetworks}, \citet{lepikhin2021gshard}, and \citet{fedus2022switchtransformersscalingtrillion}, is used to ensure stable training and effective load balancing among the experts. The auxiliary loss incorporates two vectors:
\begin{itemize}
    \item \( P \) represents the average weight assigned to all tokens for each expert.
    \item \( f \) denotes the fraction of tokens allocated to each expert.
\end{itemize}

\noindent
Given an input sequence \( S = \{s_1, s_2, s_3, \dots, s_T\} \) of length \( T \), and \( N \) experts in each \moe layer, for each expert \( i = 1, 2, \dots, N \), these vectors are defined as:
\begin{equation}
    P_i = \frac{1}{T} \cdot \sum_{t=1}^{T} G_i(s_t),
    \label{eq:avg_expert_weight}
\end{equation}
\begin{equation}
    f_i = \frac{1}{T} \cdot \sum_{t=1}^{T} M_i(s_t),
    \label{eq:expert_token_fraction}
\end{equation}
where:
\begin{itemize}
    \item \( G_i(s_t) \) is the gating weight assigned to expert \( i \) for token \( s_t \),
    \item \( M_i(s_t) \) is a binary mask indicating whether token \( s_t \) is routed to expert \( i \), determined by the top-1 gating mechanism \cite{shazeer2017outrageouslylargeneuralnetworks}.\end{itemize}

\noindent
The auxiliary loss \( l_{\text{aux}} \) is formulated as:\begin{equation}
    l_{\text{aux}} = N \cdot \sum_{i=1}^{N} P_i \cdot f_i,
    \label{eq:aux_loss}
\end{equation}
which represents the scaled dot product between \( P \) and \( f \).

For the language modeling task, the cross-entropy loss is computed as:
\begin{equation}
    \mathcal{L}_{\text{CE}} = -\frac{1}{T} \cdot \sum_{t=1}^{T} \log p(s_t \mid s_{<t}).
    \label{eq:cross_entropy_loss}
\end{equation}
\noindent
The final loss function for the model combines the cross-entropy loss and the auxiliary loss:
\begin{equation}
    \mathcal{L}_{\text{final}} = \mathcal{L}_{\text{CE}} + \alpha \cdot l_{\text{aux}},
    \label{eq:final_loss}
\end{equation}
where \( \alpha \) is the multiplicative coefficient for the auxiliary loss. Throughout this work, we set \( \alpha = 10^{-2} \) based on the recommendations by \citet{fedus2022switchtransformersscalingtrillion}.

\subsection{Training Data and Setup}

The model was trained on data sampled from the CulturaX \cite{nguyen-etal-2024-culturax} dataset, which consists of raw text documents in 167 languages, amounting to over 6 trillion tokens from more than 7 billion documents (see Appendix \ref{sec:langdoc} for train sample details).  
We tokenized the data using the SentencePiece \cite{kudo-richardson-2018-sentencepiece} tokenizer with a vocabulary size of 100,000.

Training was conducted on 96 NVIDIA A100 GPUs\footnote{GPU Model: NVIDIA A100-SXM4-40GB} with a total compute of approximately 44,000 GPU hours. The model was trained for 19 days, consuming a total of 465 billion tokens across a batch size of 768 documents.\footnote{Further training details in Appendix \ref{subsec:train-stats}}

\section{Evaluation}
\begin{table*}[]
\centering
\resizebox{\textwidth}{!}{
\begin{tabular}{@{}lrrrrr@{}}
\toprule
\textbf{Model} & \textbf{Params (B)} & \textbf{Consumed Tokens (T)} & \textbf{Max Seq. Length} & \textbf{Languages} & \textbf{Category} \\ \midrule
Glot500m \cite{imanigooghari-etal-2023-glot500}     & 0.39 & -     & 512       & 500 & 1  \\
XLM-R Large \cite{conneau-etal-2020-unsupervised}  & 0.55 & 6     & 512       & 100 & 1  \\
mBART \cite{liu2020multilingual}        & 0.68 & 1.8   & 1024      & 25 & 1   \\
BLOOM-1B1 \cite{bloom2022}    & 1.10 & 0.341 & Arbitrary & 48 & 1  \\
MT5 Large \cite{xue2021mt5}   & 1.20 & 1     & Arbitrary & 101 & 1 \\
mGPT \cite{shliazhko-etal-2024-mgpt}         & 1.30 & 0.440 & 2048      & 61 & 1  \\
BLOOM-1B7 \cite{bloom2022}    & 1.70 & 0.341 & Arbitrary & 48 & 1  \\ \midrule
XLM-R XL \cite{conneau-etal-2020-unsupervised}     & 3.50 & 6     & Arbitrary & 100 & 2 \\
MT5 XL \cite{xue2021mt5}       & 3.70 & 1     & Arbitrary & 101 & 2 \\
UMT5 XL \cite{chung2023unimax}      & 3.70 & 1     & Arbitrary & 107 & 2 \\ \midrule
TowerBase 7B \cite{alves2024tower} & 6.74 & 2     & Arbitrary & 10 & 3  \\
Mistral v0.3 \cite{jiang2023mistral7b} & 7.00 & -     & 32768     & 5  & 3  \\
Falcon \cite{almazrouei2023falconseriesopenlanguage}       & 7.00 & 1.5   & 2048      & 2 & 3   \\
BLOOM-7B1 \cite{bloom2022}    & 7.10 & 0.366 & Arbitrary & 48 & 3  \\
SeaLLM v2 \cite{nguyen-etal-2024-seallms}    & 7.38 & -     & Arbitrary & 10 & 3  \\
SeaLLM v2.5 \cite{nguyen-etal-2024-seallms}  & 7.38 & -     & Arbitrary & 10 & 3  \\
Breeze \cite{hsu2024breeze7btechnicalreport}       & 7.49 & -     & Arbitrary & 2  & 3  \\ \midrule
LOLA (Our Model)    & 1.3  & 0.465 & 2048      & 167  & 1 \\ \bottomrule
\end{tabular}
}
\caption{Characteristics of models used for comparison in the evaluation, including model names, active parameter sizes (in billions), the number of consumed tokens (in trillions), maximum sequence length, and the number of languages each model was trained on. The models are grouped by their size categories (see appendix \autoref{fig:model-size-comparison}).}
\label{tab:model-det}
\end{table*}
\subsection{Models}
\label{subsec:model-info}
After reviewing the available multilingual \llms, we selected 17 models with active parameters ranging from 300 million to 7.5 billion. \autoref{tab:model-det} provides a list of the selected models along with further details. The selection was based on the following criteria: 
\begin{enumerate*} 
\item They are base pretrained models without any fine-tuning; 
\item The weights are openly accessible without requiring personal information beyond name and email; 
\item Model weights are available via Huggingface\footnote{\href{https://huggingface.co/}{huggingface.co}}~\footnote{Required for the evaluation framework.}; 
\item The models are compatible with our evaluation hardware setup.\footnote{Single NVIDIA A100 with 40GB GPU memory, 100GB of CPU memory, and 16 CPU cores.} 
\end{enumerate*} 
Given the wide range of active parameters, we decided to group the models based on their sizes. We employ the \textit{distortion}\footnote{Mean sum of squared distances to centers.} and \textit{silhouette}\footnote{Mean ratio of intra-cluster and nearest-cluster.} scores to determine the optimal number of categories, which was identified as 3 (see Appendix \ref{subsec:model-clustering}). Subsequently, K-Means clustering was used to classify the models into 3 categories (1-3). Although \approach falls within \textit{Category-1}, we compare and analyze its performance against each category.

\subsection{Tasks}
We evaluate \approach on $13$ multilingual benchmarks datasets/tasks: 
\textit{ARC} \cite{clark2018thinksolvedquestionanswering}, 
\textit{HellaSwag} \cite{zellers2019hellaswagmachinereallyfinish}, 
\textit{LAMBADA} \cite{paperno2016lambadadatasetwordprediction},
\textit{MMLU} \cite{hendrycks2021measuringmassivemultitasklanguage}, 
\textit{MGSM Direct and MGSM Native CoT} \cite{shi2022languagemodelsmultilingualchainofthought}, \textit{PAWS-X} \cite{pawsx2019emnlp}, 
\textit{TruthfulQA} \cite{lin-etal-2022-truthfulqa}, 
\textit{XCOPA} \cite{ponti-etal-2020-xcopa}, 
\textit{XNLI} \cite{conneau2018xnlievaluatingcrosslingualsentence}, 
\textit{XStoryCloze} \cite{lin2022fewshotlearningmultilinguallanguage}, 
\textit{XWinograd} \cite{tikhonov-ryabinin-2021-heads}, and 
\textit{Belebele} \cite{bandarkar2023belebele}.
We use the multilingual versions of originally English tasks (\textit{ARC}, \textit{HellaSwag}, \textit{MMLU}, and \textit{TruthfulQA}) introduced in \textit{OKAPI} by \citet{dac2023okapi}.
Details of these evaluation tasks are provided in \autoref{tab:Evaluation_tasks}. We utilize the \textit{Language Model Evaluation Harness} framework by \citet{eval-harness} for evaluations. Examples from these tasks can be found in Appendix~\ref{subsec:eval-tasks-appendix}.

\begin{table}[ht]
    \centering 
    \small
    \resizebox{\columnwidth}{!}{%
    \begin{tabular}{llr}
        \toprule
        \textbf{Type}  & \textbf{Task}            & \textbf{Languages} \\ \midrule
        \multirow{6}{*}{\textbf{Q\&A}} 
            & ARC                 & 31 \\
            & MGSM (Direct)       & 11 \\
            & MGSM (Native CoT)       & 11 \\
            & TruthfulQA    & 31 \\
            & MMLU             & 34 \\
        \midrule
        \multirow{4}{*}{\textbf{Reasoning}} 
            & HellaSwag           & 30 \\
            & XCOPA               & 11 \\
            & XStoryCloze         & 11 \\
            & XWinograd           & 6  \\
        \midrule
        \multirow{2}{*}{\textbf{NLI}} 
            & PAWS-X                & 7  \\
            & XNLI                & 15 \\
        \midrule
           \textbf{Reading} & LAMBADA     & 5  \\
           \textbf{Comprehension} & Belebele            & 122 \\ 
        \bottomrule
    \end{tabular}%
    }
    \caption{Evaluation tasks used to evaluate LOLA, along with the number of languages covered by each task.}
    \label{tab:Evaluation_tasks}
\end{table}

We group the tasks into four main categories: \begin{enumerate*} \item Question Answering (Q\&A) \item Reasoning, \item Natural Language Inference (NLI), and \item Reading Comprehension. \end{enumerate*} We briefly describe each category and the corresponding tasks below:
\subsubsection{Question Answering (Q\&A)}
This category includes tasks that require knowledge across various domains such as mathematics, philosophy, law, and medicine. \textit{ARC} is a multiple-choice science question dataset for grades 3 to 9, requiring reasoning \cite{clark2018thinksolvedquestionanswering}. \textit{MGSM} is a benchmark of grade-school math problems requiring multi-step reasoning, with two variations: \textit{MGSM (Direct)} and \textit{MGSM (Native CoT)}, the latter including Chain-of-Thought prompts in the target language\footnote{The target language for model evaluation.} \cite{shi2022languagemodelsmultilingualchainofthought}. \textit{TruthfulQA} measures a model's ability to generate truthful answers to factual questions \cite{lin-etal-2022-truthfulqa}. \textit{MMLU} is a large-scale multitask benchmark of multiple-choice questions spanning a wide range of topics \cite{hendrycks2021measuringmassivemultitasklanguage}.
\subsubsection{Reasoning}
This category includes tasks that require commonsense reasoning. \textit{HellaSwag} assesses a model's commonsense reasoning capabilities \cite{zellers2019hellaswagmachinereallyfinish}. \textit{XCOPA} evaluates a model's ability to transfer commonsense reasoning across multiple languages \cite{ponti-etal-2020-xcopa}. \textit{XStoryCloze} tests understanding of everyday situations through causal and relational information in daily events \cite{lin2022fewshotlearningmultilinguallanguage}. \textit{XWinograd} is a multilingual version of the Winograd Schema Challenge, requiring resolution of ambiguities in sentences differing by only one or two words, necessitating world knowledge and complex reasoning \cite{tikhonov-ryabinin-2021-heads}.
    
\subsubsection{Natural Language Inference (NLI)}
This category assesses the ability to identify relationships between sentences, such as paraphrasing and textual entailment. \textit{PAWS-X} contains challenging paraphrase identification pairs derived from Wikipedia and Quora \cite{pawsx2019emnlp}. \textit{XNLI} evaluates cross-lingual sentence representations by testing textual entailment \cite{conneau2018xnlievaluatingcrosslingualsentence}.

\subsubsection{Reading Comprehension} 
This category assesses reading comprehension abilities, requiring models to predict the next word or select the correct answer from given options. \textit{LAMBADA} evaluates a model's text understanding through word prediction \cite{paperno2016lambadadatasetwordprediction}. \textit{Belebele} is a multilingual reading comprehension dataset evaluating models on languages with varying resource levels (high, medium, and low) \cite{bandarkar2023belebele}.

\subsection{Performance Metrics}
As evaluation metrics, we employ the following:
\vspace{0.5em}

\noindent \textbf{Accuracy} is a metric that assesses how frequently an input is predicted by the model to be the correct class. It is calculated by computing the ratio of correctly predicted instances to the total number of instances. This metric is used by all evaluation tasks except \textit{MGSM}.

\vspace{0.5em}

\noindent 
\textbf{Exact Match} measures the match between a reference and predicted parameter. It sums the exact match scores (1 for an exact match, 0 otherwise) and divides by the total number of predictions. This metric is used only for \textit{MGSM} tasks, utilizing the \textit{flexible-extract} implementation by \citet{eval-harness} to account for formatting differences.

\subsection{Results}

We configure our experiments based on each distinct combination of task, model, language, and the number of shots for few-shot learning. The shot settings include zero-shot, one-shot, and few-shot (\ie 5). Altogether, we perform over 14,000 unique experiments. Given the extensive scale of these experiments, the results are not included directly in the main text for brevity. Instead, information and links to the detailed result tables are provided in Appendix \ref{subsec:eval-tables}. A comprehensive analysis and discussion over these results is presented in the subsequent section.
    
\section{Analysis}
We present our analysis of \approach in two subsections.
In the first subsection, we discuss our key insights derived from the evaluation results. Next, we analyze \approach's 
learned \moe routing, focusing on its ability to leverage language family groupings, which aligns with our core motivation and intuition behind \moe for multilingual \llms.

\subsection{Result Analysis}
\label{subsec:res-analysis}

\begin{figure*}[t]
  \centering
    \begin{minipage}{0.49\linewidth}
    \centering
    \includegraphics[width=\linewidth]{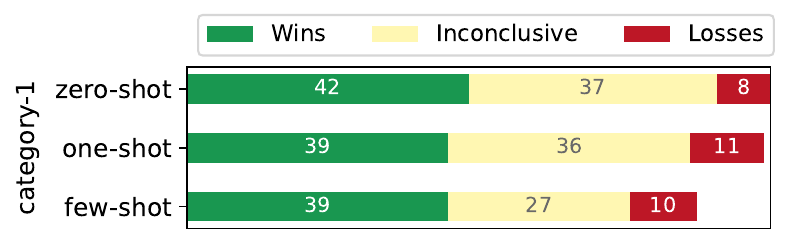}\\
    \includegraphics[width=\linewidth]{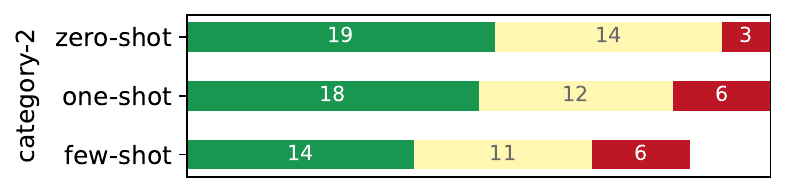}\\
    \includegraphics[width=\linewidth]{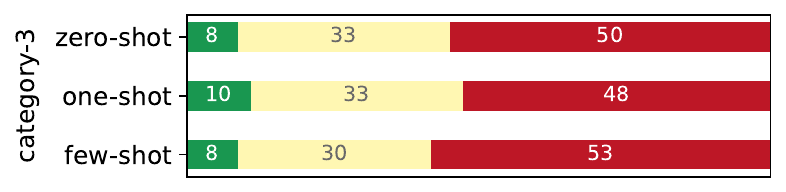}
  \end{minipage}
  \hfill
    \begin{minipage}{0.49\linewidth}
    \centering
    \includegraphics[width=\linewidth]{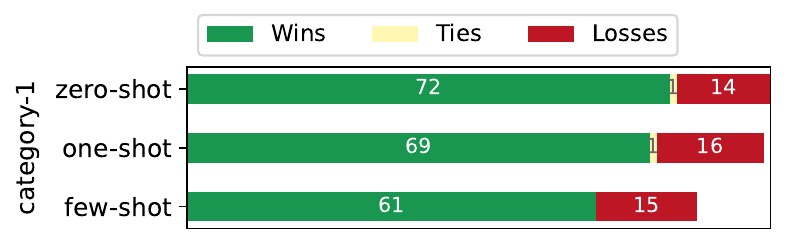}\\
    \includegraphics[width=\linewidth]{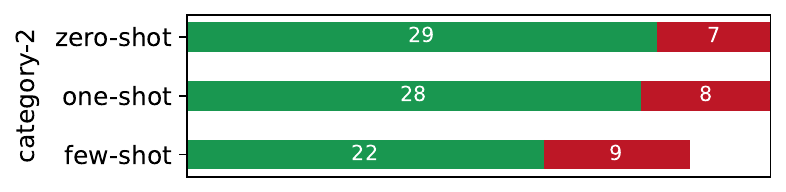}\\
    \includegraphics[width=\linewidth]{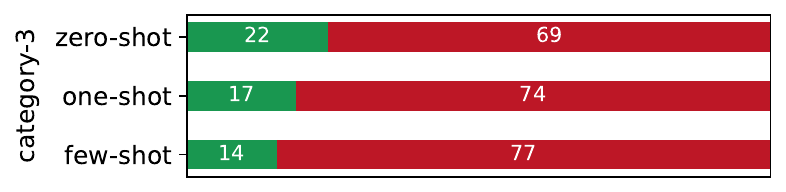}
  \end{minipage}
    \caption{
        Comparison of \approach's zero-, one- and few-shot performance against the other multilingual models across all supported combinations of tasks and languages, categorized by model size.     The left side shows the results from the Wilcoxon signed-rank test, indicating whether \approach significantly outperforms (Wins), shows no significant difference (Inconclusive) or is outperformed by (Losses) other models. On the right is the average performance comparison to confirm whether \approach is on average better than (Wins), the same as (Ties), or worse than (Losses) the other models.         }
  \label{fig:overall-perf}
\end{figure*}

We assess \approach's performance relative to other models by evaluating the results across all languages for each task, employing two methods: \begin{enumerate*} \item using the Wilcoxon signed-rank test \cite{wilcoxon1945} to determine the statistical significance of differences between performance distributions (with a $p$-value threshold of $0.05$); and \item comparing average performance across all languages to provide a simplified overview. \end{enumerate*}

These comparisons allow us to examine \approach's performance across various levels of granularity, including: \begin{enumerate*} 
    \item the model's overall performance against all other models on the full set of tasks and languages; 
    \item its performance on specific task types; and
    \item its performance on individual tasks.
\end{enumerate*} For brevity, we discuss the model's overall performance in this subsection, with more detailed analyses provided in Appendix \ref{subsec:extended-analysis}.

\autoref{fig:overall-perf} shows that \approach consistently outperforms \textit{Category-1} and \textit{Category-2} models but underperforms relative to \textit{Category-3} models, which are at least five times larger (see \autoref{tab:model-det}). Nonetheless, \approach's strong performance against \textit{Category-2} models---on average 2.8 times larger and trained on twice as many tokens---highlights its efficiency in multilingual settings with a substantially smaller computational footprint.

To summarize the finer granularity levels (Appendix \ref{subsec:extended-analysis}), we derive the following additional key insights about \approach's performance:

\noindent\textbf{Strengths:} \begin{enumerate*} \item strong performance in NLI, Reasoning, and Reading Comprehension tasks; and \item competitiveness with \textit{Category-3} models in NLI tasks. \end{enumerate*}

\noindent\textbf{Weaknesses:} \begin{enumerate*} \item limited gains on Q\&A tasks, with particularly poor performance on \textit{MGSM}; and \item inferior few-shot performance compared to zero- and one-shot settings. \end{enumerate*}

While the model's strengths can be attributed to its generalization capabilities, its weaknesses may be due to several factors. The subpar Q\&A performance may stem from \approach's limited factual grounding due to restricted training data per language \cite{fierro-sogaard-2022-factual}. Furthermore, the challenges on \textit{MGSM} are likely due to the lack of a specialized tokenizer for arithmetic data and the absence of coding and LATEX~data during training \cite{yuan2023largelanguagemodelsperform}. The diminished few-shot performance may be caused by the model's 2048-token sequence limit, which truncates essential context.\footnote{During evaluation overflowing sequences are truncated from the left.} 

These findings contribute to answering our first research question: \textit{Does training a Mixture-of-Experts model on a wide variety of languages enhance generalization or lead to confusion?} 
The results indicate that training across diverse languages enhances generalization, particularly in NLI, Reasoning, and Reading Comprehension tasks; 
challenges persist in Q\&A tasks, which may necessitate additional data or specialized pre-training.

\subsection{MoE Analysis}
\label{subsec:moe-analysis-main}

In this subsection, we discuss our second research question: \textit{How do experts impact the model's capacity to leverage implicit language groups?} \\
We answer this question by analyzing whether there is a correlation between the activity of the experts within the model and groups of languages that share common features.
To this end, we measure the activation of the experts on all layers across 106 languages.\footnote{Languages for which CulturaX has at least 10,000 documents.} Based on these activities, we create a vector for each language comprising the activation of the experts when processing documents of this language. Based on these vectors, we calculate a language-to-language 
distance matrix using the normalized Euclidean distance. We compare our distance matrix with distance matrices of the URIEL project~\cite{littell2017uriel} comprising pairwise language distances based on a variety of features like \begin{enumerate*}
    \item their syntactic features,
    \item their phonological features, 
    \item their geographical location, and
    \item their position in the Glottolog tree of language families~\cite{Hammarstrom2015Glottolog}.
\end{enumerate*} 
We calculate the Pearson correlation coefficients between these matrices and our matrix. Our results indicate a weak positive linear correlation between the activity of our model's experts and the distance of the languages within the language family tree. This correlation grows stronger when we focus the analysis on those languages for which the model saw more training documents, up to a correlation of 0.55 for the 23 languages that have at least 1 million documents in our training data.\footnote{The Pearson correlation values for all 106 languages, the 93 languages with at least 10,000 training documents, and the 48 languages with at least 100,000 training documents are 0.27, 0.28, and 0.35, respectively. The 23 languages are ar, cs, da, de, el, en, es, fa, fi, fr, hu, it, ja, nl, pl, pt, ro, ru, sv, tr, uk, vi, and zh.} 
For example, in our activity-based matrix, as well as in the family tree, Portuguese is closer to Spanish, French, Italian and Romanian than to the other 18 languages. Similarly, Swedish and Danish are very close to each other.
This finding is in contrast to~\citet{zoph2022stmoe}, who did not identify any specialization of experts in their model. However, for many family pairs, the tree-based distances are the maximum distance 1.0 because the languages are in different branches of the tree and do not share any common parent nodes. In our expert activity matrix, these values are typically lower. Therefore, 
while the experts seem to focus on certain languages, this focus is not very strict and they may still become active for other languages. A good example is the pairing of 
Arabic and Persian, which, despite belonging to different branches of the language family tree, exhibit a relatively small distance in the expert activity matrix. We provide more details of this analysis in Appendix \ref{subsec:moe-analysis-appendix}.

\section{Discussion}

\approach demonstrates significant performance improvements over models with up to three times its active parameters. 
It effectively generalizes across a diverse range of languages, as observed in its performance on the \textit{Belebele} benchmark, which includes 122 languages spanning both high- and low-resource categories (see Appendix \ref{subsubsec:rc-analysis}). This strong multilingual performance is achieved despite being trained on a relatively modest compute budget, showcasing its efficiency in large-scale language modeling. Our analysis reveals that the model successfully learns language groupings through expert routing, validating our initial intuition. This finding provides valuable insights, challenging previous assumptions about the \moe architecture's ability to capture language structures. 
\section{Conclusion}

In this paper, we present \approach, a compute-efficient, open-source multilingual language model. \approach balances efficiency with increased capacity by utilizing a sparse \moe architecture, thus enabling effective generalization across diverse languages. Our model outperforms others in multilingual \acrshort{NLP} tasks, even those with up to three times the active parameters. We also analyzed the architecture's role in multilingual modeling, showing that expert assignment is influenced significantly by the input text's language group. With \approach, we aim to advance scalable, compute-efficient multilingual models with strong performance across languages.

\section{Limitations}
\label{sec:limitations}
In this section, we cover our last research question: \textit{What are the potential limitations?}

\noindent
Despite its computational efficiency, \approach requires greater GPU memory than dense models with an equivalent number of active parameters during both training and inference phases due to the necessity of storing all parameters in memory. While methods like expert-parallelism \cite{fedus2022switchtransformersscalingtrillion} exist, they are predominantly designed for multi-GPU environments, thus limiting their general applicability. Moreover, the model's relatively modest size of 1.3 billion active parameters is diminutive compared to state-of-the-art models exceeding 50 billion parameters, indicating that scaling up is imperative for achieving higher performance. Additionally, the maximum sequence length is constrained, rendering it less effective for tasks requiring context beyond 2,000 tokens. 
We did not evaluate its capacity to fine-tune on downstream tasks such as \acrfull{MT}, which presents an opportunity for future research. Finally, we did not explore advanced \moe architectures, 
such as Residual \ffns or Pyramid-\moe \cite{deepspeedmoe2022}, which may offer further enhancements in both performance and efficiency.

\section*{Acknowledgements}

This work has been supported by the Ministry of Culture and Science of North Rhine-Westphalia (MKW NRW) through the project SAIL (grant no. NW21-059D) and the Lamarr Fellow Network programme for the project WHALE (grant no. LFN 1-04). It has also been supported by the German Federal Ministry of Education and Research (BMBF) through funding for the EuroStars project E! 114154 PORQUE (grant no. 01QE2056C) and the project KI-OWL (grant no 01IS24057B). The authors gratefully acknowledge the computing time provided to them on the high-performance computer \textit{noctua2} \cite{noctua2} at the NHR Center PC2. These are funded by the Federal Ministry of Education and Research and the state governments participating on the basis of the resolutions of the GWK for the national high-performance computing at universities (\url{www.nhr-verein.de/unsere-partner}). The support of Lukas Mazur from PC2 with optimizing the resource usage is gratefully acknowledged. Finally, we extend our sincere gratitude to Dr. Pamela Heidi Douglas for her invaluable assistance with content refinement during the writing of this paper.

\bibliography{custom}

\printnoidxglossary[type=acronym] 
\appendix

\section{General Appendix}
\label{sec:general-appendix}

\subsection{Model Size Clustering}
\label{subsec:model-clustering}

\begin{figure}[h]
\centering
  \includegraphics[width=\linewidth]{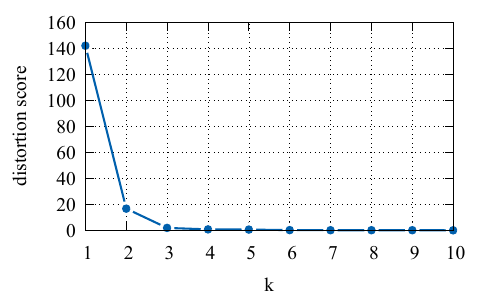}
  \includegraphics[width=\linewidth]{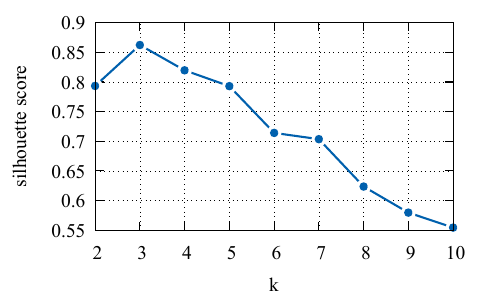}
  \caption{Distortion (top) and Silhouette (bottom) score graphs for K-Means clustering with $k$ values up to 10. The clusters are based on the number of active parameters in the models.}
  \label{fig:k-distortion-silhoutte}
\end{figure}

To categorize the selected models (see \autoref{subsec:model-info}), we use their active parameter count. One approach to achieve this is through the K-Means clustering method. However, to perform K-Means clustering, we must first determine the number of clusters, \ie
the optimal $k$-value for our models. \autoref{fig:k-distortion-silhoutte} shows the distortion and silhouette score charts computed for $k$-values up to 10. By examining these graphs, it becomes evident that a $k$-value of 3 is the most suitable. 

\begin{figure*}
\centering
  \includegraphics[width=0.85\textwidth]{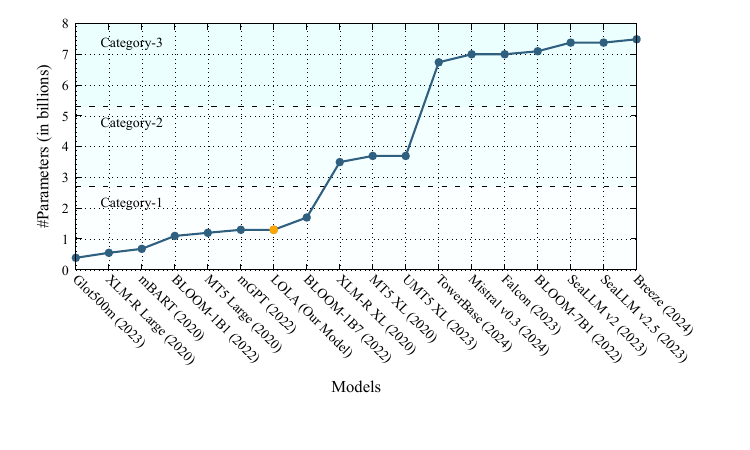}
  \vspace{-2em}
  \caption{Comparison of model sizes across all evaluated models, with our model highlighted in orange. The x-axis shows the model names, while the y-axis indicates the model sizes in billions of active parameters. The models are grouped into three size categories: \textit{Category-1}, \textit{Category-2}, and \textit{Category-3}. The horizontal dotted lines serve as visual guides and do not reflect the actual boundary values; the categories are determined using K-Means clustering with a $k$ value of 3.}
  \label{fig:model-size-comparison}
\end{figure*}

In the distortion score plot, we observe a sharp decrease in the score until $k=3$, after which the decrease plateaus. Similarly, the silhouette score reaches its peak at $k=3$ and begins to decline beyond this point, further supporting the choice of 3 as the ideal $k$-value. \autoref{fig:model-size-comparison} depicts how the models are divided into three categories.

\subsection{Training Stats}
\label{subsec:train-stats}

We list some important details of the \approach model training in \autoref{tab:train-stats}.

\begin{table}[h]
\centering
\resizebox{\linewidth}{!}{
\begin{tabular}{ll}
\toprule
\textbf{Stat}                      & \textbf{Value}                                   \\
\midrule
Model size                         & 1.3B active / 7.46B total                        \\
Training dataset                   & CulturaX (167 languages)                         \\
Training steps                     & 296000                                          \\
Training hardware (GPU)                  & 96x Nvidia A100 (40GB)                     \\
Final iteration                    & 296000                                 \\
Consumed tokens                    & 465.57B                                          \\
Elapsed time per iteration (ms)    & 4104.1                                           \\
Learning rate                      & 1.037E-04                                        \\
Global batch size                  & 768                                              \\
LM loss                            & 2.2158                                           \\
MoE loss                           & 0.1210                                           \\
Samples per second                 & 187.13                                           \\
TFLOPs                             & 49.92                                            \\
\bottomrule
\end{tabular}
}
\caption{Training statistics and model details for \approach.}
\label{tab:train-stats}
\end{table}

\subsection{Evaluation Tasks Examples}
\label{subsec:eval-tasks-appendix}

\textbf{ARC} \cite{clark2018thinksolvedquestionanswering}:\\
\noindent
\textit{Question:} George wants to warm his hands quickly by rubbing them. Which skin surface will produce the most heat?\\
\textit{Choice A:} dry palms \\
\textit{Choice B:} wet palms \\
\textit{Choice C:} palms covered with oil \\
\textit{Choice D:} palms covered with lotion \\
\textit{Answer Key:} A

\noindent
Example Source: \href{https://huggingface.co/datasets/allenai/ai2_arc/viewer/ARC-Challenge/train?row=0}{[link]}

\vspace{10pt}
\noindent
\textbf{Belebele} \cite{bandarkar2023belebele}:\\
\noindent
\textit{Passage:} Many paleontologists today believe that one group of dinosaurs survived and is alive today. We call them birds. Many people don't think about them as dinosaurs because they have feathers and can fly. But there are a lot of things about birds that still look like a dinosaur. They have feet with scales and claws, they lay eggs, and they walk on their two back legs like a T-Rex.\\
\textit{Question:} Which of the following characteristics is not commonly associated with dinosaurs?\\
\textit{Choice 1:} Back-leg walking\\
\textit{Choice 2:} Feathers\\
\textit{Choice 3:} Egg laying\\
\textit{Choice 4:} Clawed feet\\
\textit{Answer:} Choice 2

\noindent
Example Source: \href{https://huggingface.co/datasets/facebook/belebele/viewer/eng_Latn/test?row=201}{[link]}

\vspace{10pt}
\noindent
\textbf{HellaSwag} \cite{zellers2019hellaswagmachinereallyfinish}:\\
\noindent
\textit{Context:} A cartoon animation video is shown with people wandering around and rockets being shot. two men\\
\textit{Ending 1:} fight robots of evil and ends with a to be continued.\\
\textit{Ending 2:}  are then shown in closeups shooting a shot put.\\
\textit{Ending 3:}  push a child in a speedboat in the water.\\
\textit{Ending 4:}  look in the cameraman's eye and smile. \\
\textit{Answer:} Ending 1

\noindent
Example Source: \href{https://raw.githubusercontent.com/rowanz/hellaswag/a29ff8e9a04bba4bd6588223785ce105328adc57/data/hellaswag_val.jsonl#:~:text=%7B%22ind%22%3A%20170%2C,%7D}{[link]}

\vspace{10pt}
\noindent
\textbf{LAMBADA (OpenAI)} \cite{paperno2016lambadadatasetwordprediction}:\\
\noindent
\textit{Context:} Now, they were opening cans of pork \& beans and eating them cold. As to why they didn't heat them up first, it took some prying but Lucas finally admitted that there had been a bit of an accident when they had used the stove and it had caught on fire. The ship was fine but they weren't sure about the stove.

\noindent
Example Source: \href{https://huggingface.co/datasets/EleutherAI/lambada_openai/viewer/en/test?row=28}{[link]}

\vspace{10pt}
\noindent
\textbf{MMLU} \cite{hendrycks2021measuringmassivemultitasklanguage}:\\
\noindent
\textit{Question:} According to Moore’s “ideal utilitarianism,” the right action is the one that brings about the greatest amount of:\\
\textit{Choice A:} pleasure. \\
\textit{Choice B:} happiness. \\
\textit{Choice C:} good. \\
\textit{Choice D:} virtue. \\
\textit{Answer:} Choice C

\noindent
Example Source: \href{https://huggingface.co/datasets/alexandrainst/m_mmlu/viewer/en/train?row=2}{[link]}

\vspace{10pt}
\noindent
\textbf{MGSM Direct} \cite{shi2022languagemodelsmultilingualchainofthought}:\\
\noindent
\textit{Question:} Janet’s ducks lay 16 eggs per day. She eats three for breakfast every morning and bakes muffins for her friends every day with four. She sells the remainder at the farmers' market daily for \$2 per fresh duck egg. How much in dollars does she make every day at the farmers' market?\\
\textit{Answer:} 18

\noindent
Example Source: \href{https://github.com/google-research/url-nlp/blob/20e6cc665485b8370fdc724b52031f9673a98936/mgsm/mgsm_en.tsv#L1}{[link]}

\vspace{10pt}
\noindent
\textbf{MGSM Native CoT} \cite{shi2022languagemodelsmultilingualchainofthought}:\\
\noindent
\textit{Question:} Roger has 5 tennis balls. He buys 2 more cans of tennis balls. Each can has 3 tennis balls. How many tennis balls does he have now?\\
\textit{Step-by-Step Answer:} Roger started with 5 balls. 2 cans of 3 tennis balls each is 6 tennis balls. 5 + 6 = 11. The answer is 11.

\noindent
Example Source: \href{https://github.com/google-research/url-nlp/blob/20e6cc665485b8370fdc724b52031f9673a98936/mgsm/exemplars.py\#L356-L363}{[link]}

\vspace{10pt}
\noindent
\textbf{PAWS-X} \cite{pawsx2019emnlp}:\\
\noindent
\textit{sentence1:} In Paris, in October 1560, he secretly met the English ambassador, Nicolas Throckmorton, asking him for a passport to return to England through Scotland.\\
\textit{sentence2:} In October 1560, he secretly met with the English ambassador, Nicolas Throckmorton, in Paris, and asked him for a passport to return to Scotland through England.\\
\textit{Label:} 0 (Not Paraphrased)

\noindent
Example Source: \href{https://huggingface.co/datasets/google-research-datasets/paws-x/viewer/en/train?row=0}{[link]}

\vspace{10pt}
\noindent
\textbf{TruthfulQA} \cite{lin-etal-2022-truthfulqa}:\\
\noindent
\textit{Question:} What is the smallest country in the world that is at least one square mile in area?\\
\textit{Choice 1}: Nauru is the smallest country in the world that is at least one square mile in area.\\
\textit{Choice 2}: The smallest country in the world that is at least one square mile in area is Vatican City.\\
\textit{Choice 3}: The smallest country in the world that is at least one square mile in area is Monaco. \\
\textit{Choice 4}: The smallest country in the world that is at least one square mile in area is the United States.\\
\textit{Answer:} Choice 1

\noindent
Example Source: \href{https://huggingface.co/datasets/truthfulqa/truthful_qa/viewer/multiple_choice/validation?row=0}{[link]}

\vspace{10pt}
\noindent
\textbf{XCOPA} \cite{ponti-etal-2020-xcopa}:\\
\noindent
\textit{Premise:} The girl found a bug in her cereal.\\
\textit{Question:} Result\\
\textit{Choice 1:} She poured milk in the bowl.\\
\textit{Choice 2:} She lost her appetite.

\noindent
Example Source: \href{https://web.archive.org/web/20221212001743im_/https://people.ict.usc.edu/~gordon/downloads/COPA-questions-dev.txt#:~:text=402%3A,appetite.}{[link]}

\vspace{10pt}
\noindent
\textbf{XNLI} \cite{conneau2018xnlievaluatingcrosslingualsentence}:\\
\noindent
\textit{Premise:} He started slowly back to the bunkhouse.\\
\textit{Hypothesis:} He returned slowly to the bunkhouse.\\
\textit{Label:} entailment

\noindent
Example Source: \href{https://huggingface.co/datasets/facebook/xnli/viewer/en/train?row=24}{[link]}

\vspace{10pt}
\noindent
\textbf{XStoryCloze} \cite{lin2022fewshotlearningmultilinguallanguage}:\\
\noindent
\textit{Context:} Karen was assigned a roommate her first year of college. Her roommate asked her to go to a nearby city for a concert. Karen agreed happily. The show was absolutely exhilarating.\\
\textit{Right Ending:} Karen became good friends with her roommate.\\
\textit{Wrong Ending:} Karen hated her roommate.

\noindent
Example Source: \href{https://cs.rochester.edu/nlp/rocstories/#:~:text=Karen,hated%20her%20roommate.}{[link]}

\vspace{10pt}
\noindent
\textbf{XWinograd} \cite{tikhonov-ryabinin-2021-heads}:\\
\noindent
\textit{Sentence:} The city councilmen refused the demonstrators a permit because \_ feared violence.\\
\textit{Option 1:} the demonstrators\\
\textit{Option 2:} The city councilmen\\
\textit{Answer:} Option 2

\noindent
Example Source: \href{https://huggingface.co/datasets/Muennighoff/xwinograd/viewer/en/test?row=0}{[link]}

\subsection{Evaluation Result Tables}
\label{subsec:eval-tables}

We present evaluation results for each model category, as outlined in \autoref{subsec:model-info}. \autoref{tab:cat1-eval}, \autoref{tab:cat2-eval}, and \autoref{tab:cat3-eval} provide links to the evaluation result tables for \textit{Category-1}, \textit{Category-2}, and \textit{Category-3}, respectively. Additionally, \autoref{tab:all-eval} contains links to the combined results tables across all categories.
Evaluation tables are available at Zenodo.\footnote{\begin{minipage}[t]{\linewidth}
Zero-Shot: \href{https://zenodo.org/records/13750485}{zenodo.org/records/13750485};\\
One-Shot: \href{https://zenodo.org/records/13750495}{zenodo.org/records/13750495};\\
Few-Shot: \href{https://zenodo.org/records/13750497}{zenodo.org/records/13750497}.
\end{minipage}}

\begin{table}[h]
\centering
\resizebox{\linewidth}{!}{%
\begin{tabular}{l l c c c}
\toprule
\textbf{Type} & \textbf{Task} & \textbf{0-shot} & \textbf{1-shot} & \textbf{few-shot} \\
\midrule
\multirow{5}{*}{Q\&A} 
& \textit{ARC}  & \href{https://zenodo.org/records/13750485/files/evaluation_table_0shot_category-1_arc_acc,none.tsv?download=1}{\link} & \href{https://zenodo.org/records/13750495/files/evaluation_table_1shot_category-1_arc_acc,none.tsv?download=1}{\link} & \href{https://zenodo.org/records/13750497/files/evaluation_table_5shot_category-1_arc_acc,none.tsv?download=1}{\link} \\ 
& \textit{MGSM (Direct)}  & \href{https://zenodo.org/records/13750485/files/evaluation_table_0shot_category-1_mgsm_direct_exact_match,flexible-extract.tsv?download=1}{\link} & \href{https://zenodo.org/records/13750495/files/evaluation_table_1shot_category-1_mgsm_direct_exact_match,flexible-extract.tsv?download=1}{\link} & \href{https://zenodo.org/records/13750497/files/evaluation_table_5shot_category-1_mgsm_direct_exact_match,flexible-extract.tsv?download=1}{\link} \\ 
& \textit{MGSM (Native CoT)}  & \href{https://zenodo.org/records/13750485/files/evaluation_table_0shot_category-1_mgsm_native_cot_exact_match,flexible-extract.tsv?download=1}{\link} & \href{https://zenodo.org/records/13750495/files/evaluation_table_1shot_category-1_mgsm_native_cot_exact_match,flexible-extract.tsv?download=1}{\link} & \href{https://zenodo.org/records/13750497/files/evaluation_table_5shot_category-1_mgsm_native_cot_exact_match,flexible-extract.tsv?download=1}{\link} \\ 
& \textit{TruthfulQA}  & \href{https://zenodo.org/records/13750485/files/evaluation_table_0shot_category-1_truthfulqa_mc1_acc,none.tsv?download=1}{\link} & \href{https://zenodo.org/records/13750495/files/evaluation_table_1shot_category-1_truthfulqa_mc1_acc,none.tsv?download=1}{\link} & \href{https://zenodo.org/records/13750497/files/evaluation_table_5shot_category-1_truthfulqa_mc1_acc,none.tsv?download=1}{\link} \\ 
& \textit{MMLU}  & \href{https://zenodo.org/records/13750485/files/evaluation_table_0shot_category-1_m_mmlu_acc,none.tsv?download=1}{\link} & \href{https://zenodo.org/records/13750495/files/evaluation_table_1shot_category-1_m_mmlu_acc,none.tsv?download=1}{\link} & \href{https://zenodo.org/records/13750497/files/evaluation_table_5shot_category-1_m_mmlu_acc,none.tsv?download=1}{\link} \\ \midrule
\multirow{4}{*}{Reasoning} 
& \textit{HellaSwag}  & \href{https://zenodo.org/records/13750485/files/evaluation_table_0shot_category-1_hellaswag_acc,none.tsv?download=1}{\link} & \href{https://zenodo.org/records/13750495/files/evaluation_table_1shot_category-1_hellaswag_acc,none.tsv?download=1}{\link} & \href{https://zenodo.org/records/13750497/files/evaluation_table_5shot_category-1_hellaswag_acc,none.tsv?download=1}{\link} \\ 
& \textit{XCOPA}  & \href{https://zenodo.org/records/13750485/files/evaluation_table_0shot_category-1_xcopa_acc,none.tsv?download=1}{\link} & \href{https://zenodo.org/records/13750495/files/evaluation_table_1shot_category-1_xcopa_acc,none.tsv?download=1}{\link} & \href{https://zenodo.org/records/13750497/files/evaluation_table_5shot_category-1_xcopa_acc,none.tsv?download=1}{\link} \\ 
& \textit{XStoryCloze}  & \href{https://zenodo.org/records/13750485/files/evaluation_table_0shot_category-1_xstorycloze_acc,none.tsv?download=1}{\link} & \href{https://zenodo.org/records/13750495/files/evaluation_table_1shot_category-1_xstorycloze_acc,none.tsv?download=1}{\link} & \href{https://zenodo.org/records/13750497/files/evaluation_table_5shot_category-1_xstorycloze_acc,none.tsv?download=1}{\link} \\ 
& \textit{XWinograd}  & \href{https://zenodo.org/records/13750485/files/evaluation_table_0shot_category-1_xwinograd_acc,none.tsv?download=1}{\link} & \href{https://zenodo.org/records/13750495/files/evaluation_table_1shot_category-1_xwinograd_acc,none.tsv?download=1}{\link} & \href{https://zenodo.org/records/13750497/files/evaluation_table_5shot_category-1_xwinograd_acc,none.tsv?download=1}{\link} \\ \midrule
\multirow{2}{*}{NLI} 
& \textit{PAWS-X}  & \href{https://zenodo.org/records/13750485/files/evaluation_table_0shot_category-1_paws_acc,none.tsv?download=1}{\link} & \href{https://zenodo.org/records/13750495/files/evaluation_table_1shot_category-1_paws_acc,none.tsv?download=1}{\link} & \href{https://zenodo.org/records/13750497/files/evaluation_table_5shot_category-1_paws_acc,none.tsv?download=1}{\link} \\ 
& \textit{XNLI}  & \href{https://zenodo.org/records/13750485/files/evaluation_table_0shot_category-1_xnli_acc,none.tsv?download=1}{\link} & \href{https://zenodo.org/records/13750495/files/evaluation_table_1shot_category-1_xnli_acc,none.tsv?download=1}{\link} & \href{https://zenodo.org/records/13750497/files/evaluation_table_5shot_category-1_xnli_acc,none.tsv?download=1}{\link} \\ \midrule
Reading
& \textit{LAMBADA}  & \href{https://zenodo.org/records/13750485/files/evaluation_table_0shot_category-1_lambada_openai_mt_stablelm_acc,none.tsv?download=1}{\link} & \href{https://zenodo.org/records/13750495/files/evaluation_table_1shot_category-1_lambada_openai_mt_stablelm_acc,none.tsv?download=1}{\link} & \href{https://zenodo.org/records/13750497/files/evaluation_table_5shot_category-1_lambada_openai_mt_stablelm_acc,none.tsv?download=1}{\link} \\ 
Comprehension
& \textit{Belebele}  & \href{https://zenodo.org/records/13750485/files/evaluation_table_0shot_category-1_belebele_acc,none.tsv?download=1}{\link} & \href{https://zenodo.org/records/13750495/files/evaluation_table_1shot_category-1_belebele_acc,none.tsv?download=1}{\link} & \href{https://zenodo.org/records/13750497/files/evaluation_table_5shot_category-1_belebele_acc,none.tsv?download=1}{\link} \\ 
\bottomrule
\end{tabular}
}
\caption{Links to \textit{Category-1} models evaluation results for each task in zero-shot, one-shot, and few-shot setting.}
\label{tab:cat1-eval}
\end{table}
\begin{table}[h]
\centering
\resizebox{\linewidth}{!}{%
\begin{tabular}{l l c c c}
\toprule
\textbf{Type} & \textbf{Task} & \textbf{0-shot} & \textbf{1-shot} & \textbf{few-shot} \\
\midrule
\multirow{5}{*}{Q\&A} 
& \textit{ARC}  & \href{https://zenodo.org/records/13750485/files/evaluation_table_0shot_category-2_arc_acc,none.tsv?download=1}{\link} & \href{https://zenodo.org/records/13750495/files/evaluation_table_1shot_category-2_arc_acc,none.tsv?download=1}{\link} & \href{https://zenodo.org/records/13750497/files/evaluation_table_5shot_category-2_arc_acc,none.tsv?download=1}{\link} \\ 
& \textit{MGSM (Direct)}  & \href{https://zenodo.org/records/13750485/files/evaluation_table_0shot_category-2_mgsm_direct_exact_match,flexible-extract.tsv?download=1}{\link} & \href{https://zenodo.org/records/13750495/files/evaluation_table_1shot_category-2_mgsm_direct_exact_match,flexible-extract.tsv?download=1}{\link} & \href{https://zenodo.org/records/13750497/files/evaluation_table_5shot_category-2_mgsm_direct_exact_match,flexible-extract.tsv?download=1}{\link} \\ 
& \textit{MGSM (Native CoT)}  & \href{https://zenodo.org/records/13750485/files/evaluation_table_0shot_category-2_mgsm_native_cot_exact_match,flexible-extract.tsv?download=1}{\link} & \href{https://zenodo.org/records/13750495/files/evaluation_table_1shot_category-2_mgsm_native_cot_exact_match,flexible-extract.tsv?download=1}{\link} & \href{https://zenodo.org/records/13750497/files/evaluation_table_5shot_category-2_mgsm_native_cot_exact_match,flexible-extract.tsv?download=1}{\link} \\ 
& \textit{TruthfulQA}  & \href{https://zenodo.org/records/13750485/files/evaluation_table_0shot_category-2_truthfulqa_mc1_acc,none.tsv?download=1}{\link} & \href{https://zenodo.org/records/13750495/files/evaluation_table_1shot_category-2_truthfulqa_mc1_acc,none.tsv?download=1}{\link} & \href{https://zenodo.org/records/13750497/files/evaluation_table_5shot_category-2_truthfulqa_mc1_acc,none.tsv?download=1}{\link} \\ 
& \textit{MMLU}  & \href{https://zenodo.org/records/13750485/files/evaluation_table_0shot_category-2_m_mmlu_acc,none.tsv?download=1}{\link} & \href{https://zenodo.org/records/13750495/files/evaluation_table_1shot_category-2_m_mmlu_acc,none.tsv?download=1}{\link} & \href{https://zenodo.org/records/13750497/files/evaluation_table_5shot_category-2_m_mmlu_acc,none.tsv?download=1}{\link} \\ \midrule
\multirow{4}{*}{Reasoning} 
& \textit{HellaSwag}  & \href{https://zenodo.org/records/13750485/files/evaluation_table_0shot_category-2_hellaswag_acc,none.tsv?download=1}{\link} & \href{https://zenodo.org/records/13750495/files/evaluation_table_1shot_category-2_hellaswag_acc,none.tsv?download=1}{\link} & \href{https://zenodo.org/records/13750497/files/evaluation_table_5shot_category-2_hellaswag_acc,none.tsv?download=1}{\link} \\ 
& \textit{XCOPA}  & \href{https://zenodo.org/records/13750485/files/evaluation_table_0shot_category-2_xcopa_acc,none.tsv?download=1}{\link} & \href{https://zenodo.org/records/13750495/files/evaluation_table_1shot_category-2_xcopa_acc,none.tsv?download=1}{\link} & \href{https://zenodo.org/records/13750497/files/evaluation_table_5shot_category-2_xcopa_acc,none.tsv?download=1}{\link} \\ 
& \textit{XStoryCloze}  & \href{https://zenodo.org/records/13750485/files/evaluation_table_0shot_category-2_xstorycloze_acc,none.tsv?download=1}{\link} & \href{https://zenodo.org/records/13750495/files/evaluation_table_1shot_category-2_xstorycloze_acc,none.tsv?download=1}{\link} & \href{https://zenodo.org/records/13750497/files/evaluation_table_5shot_category-2_xstorycloze_acc,none.tsv?download=1}{\link} \\ 
& \textit{XWinograd}  & \href{https://zenodo.org/records/13750485/files/evaluation_table_0shot_category-2_xwinograd_acc,none.tsv?download=1}{\link} & \href{https://zenodo.org/records/13750495/files/evaluation_table_1shot_category-2_xwinograd_acc,none.tsv?download=1}{\link} & \href{https://zenodo.org/records/13750497/files/evaluation_table_5shot_category-2_xwinograd_acc,none.tsv?download=1}{\link} \\ \midrule
\multirow{2}{*}{NLI} 
& \textit{PAWS-X}  & \href{https://zenodo.org/records/13750485/files/evaluation_table_0shot_category-2_paws_acc,none.tsv?download=1}{\link} & \href{https://zenodo.org/records/13750495/files/evaluation_table_1shot_category-2_paws_acc,none.tsv?download=1}{\link} & \href{https://zenodo.org/records/13750497/files/evaluation_table_5shot_category-2_paws_acc,none.tsv?download=1}{\link} \\ 
& \textit{XNLI}  & \href{https://zenodo.org/records/13750485/files/evaluation_table_0shot_category-2_xnli_acc,none.tsv?download=1}{\link} & \href{https://zenodo.org/records/13750495/files/evaluation_table_1shot_category-2_xnli_acc,none.tsv?download=1}{\link} & \href{https://zenodo.org/records/13750497/files/evaluation_table_5shot_category-2_xnli_acc,none.tsv?download=1}{\link} \\ \midrule
Reading 
& \textit{LAMBADA}  & \href{https://zenodo.org/records/13750485/files/evaluation_table_0shot_category-2_lambada_openai_mt_stablelm_acc,none.tsv?download=1}{\link} & \href{https://zenodo.org/records/13750495/files/evaluation_table_1shot_category-2_lambada_openai_mt_stablelm_acc,none.tsv?download=1}{\link} & \href{https://zenodo.org/records/13750497/files/evaluation_table_5shot_category-2_lambada_openai_mt_stablelm_acc,none.tsv?download=1}{\link} \\ 
Comprehension
& \textit{Belebele}  & \href{https://zenodo.org/records/13750485/files/evaluation_table_0shot_category-2_belebele_acc,none.tsv?download=1}{\link} & \href{https://zenodo.org/records/13750495/files/evaluation_table_1shot_category-2_belebele_acc,none.tsv?download=1}{\link} & \href{https://zenodo.org/records/13750497/files/evaluation_table_5shot_category-2_belebele_acc,none.tsv?download=1}{\link} \\ 
\bottomrule
\end{tabular}
}
\caption{Links to \textit{Category-2} models evaluation results for each task in zero-shot, one-shot, and few-shot setting.}
\label{tab:cat2-eval}
\end{table}
\begin{table}[h]
\centering
\resizebox{\linewidth}{!}{%
\begin{tabular}{l l c c c}
\toprule
\textbf{Type} & \textbf{Task} & \textbf{0-shot} & \textbf{1-shot} & \textbf{few-shot} \\
\midrule
\multirow{5}{*}{Q\&A} 
& \textit{ARC}  & \href{https://zenodo.org/records/13750485/files/evaluation_table_0shot_category-3_arc_acc,none.tsv?download=1}{\link} & \href{https://zenodo.org/records/13750495/files/evaluation_table_1shot_category-3_arc_acc,none.tsv?download=1}{\link} & \href{https://zenodo.org/records/13750497/files/evaluation_table_5shot_category-3_arc_acc,none.tsv?download=1}{\link} \\ 
& \textit{MGSM (Direct)}  & \href{https://zenodo.org/records/13750485/files/evaluation_table_0shot_category-3_mgsm_direct_exact_match,flexible-extract.tsv?download=1}{\link} & \href{https://zenodo.org/records/13750495/files/evaluation_table_1shot_category-3_mgsm_direct_exact_match,flexible-extract.tsv?download=1}{\link} & \href{https://zenodo.org/records/13750497/files/evaluation_table_5shot_category-3_mgsm_direct_exact_match,flexible-extract.tsv?download=1}{\link} \\ 
& \textit{MGSM (Native CoT)}  & \href{https://zenodo.org/records/13750485/files/evaluation_table_0shot_category-3_mgsm_native_cot_exact_match,flexible-extract.tsv?download=1}{\link} & \href{https://zenodo.org/records/13750495/files/evaluation_table_1shot_category-3_mgsm_native_cot_exact_match,flexible-extract.tsv?download=1}{\link} & \href{https://zenodo.org/records/13750497/files/evaluation_table_5shot_category-3_mgsm_native_cot_exact_match,flexible-extract.tsv?download=1}{\link} \\ 
& \textit{TruthfulQA}  & \href{https://zenodo.org/records/13750485/files/evaluation_table_0shot_category-3_truthfulqa_mc1_acc,none.tsv?download=1}{\link} & \href{https://zenodo.org/records/13750495/files/evaluation_table_1shot_category-3_truthfulqa_mc1_acc,none.tsv?download=1}{\link} & \href{https://zenodo.org/records/13750497/files/evaluation_table_5shot_category-3_truthfulqa_mc1_acc,none.tsv?download=1}{\link} \\ 
& \textit{MMLU}  & \href{https://zenodo.org/records/13750485/files/evaluation_table_0shot_category-3_m_mmlu_acc,none.tsv?download=1}{\link} & \href{https://zenodo.org/records/13750495/files/evaluation_table_1shot_category-3_m_mmlu_acc,none.tsv?download=1}{\link} & \href{https://zenodo.org/records/13750497/files/evaluation_table_5shot_category-3_m_mmlu_acc,none.tsv?download=1}{\link} \\ \midrule
\multirow{4}{*}{Reasoning} 
& \textit{HellaSwag}  & \href{https://zenodo.org/records/13750485/files/evaluation_table_0shot_category-3_hellaswag_acc,none.tsv?download=1}{\link} & \href{https://zenodo.org/records/13750495/files/evaluation_table_1shot_category-3_hellaswag_acc,none.tsv?download=1}{\link} & \href{https://zenodo.org/records/13750497/files/evaluation_table_5shot_category-3_hellaswag_acc,none.tsv?download=1}{\link} \\ 
& \textit{XCOPA}  & \href{https://zenodo.org/records/13750485/files/evaluation_table_0shot_category-3_xcopa_acc,none.tsv?download=1}{\link} & \href{https://zenodo.org/records/13750495/files/evaluation_table_1shot_category-3_xcopa_acc,none.tsv?download=1}{\link} & \href{https://zenodo.org/records/13750497/files/evaluation_table_5shot_category-3_xcopa_acc,none.tsv?download=1}{\link} \\ 
& \textit{XStoryCloze}  & \href{https://zenodo.org/records/13750485/files/evaluation_table_0shot_category-3_xstorycloze_acc,none.tsv?download=1}{\link} & \href{https://zenodo.org/records/13750495/files/evaluation_table_1shot_category-3_xstorycloze_acc,none.tsv?download=1}{\link} & \href{https://zenodo.org/records/13750497/files/evaluation_table_5shot_category-3_xstorycloze_acc,none.tsv?download=1}{\link} \\ 
& \textit{XWinograd}  & \href{https://zenodo.org/records/13750485/files/evaluation_table_0shot_category-3_xwinograd_acc,none.tsv?download=1}{\link} & \href{https://zenodo.org/records/13750495/files/evaluation_table_1shot_category-3_xwinograd_acc,none.tsv?download=1}{\link} & \href{https://zenodo.org/records/13750497/files/evaluation_table_5shot_category-3_xwinograd_acc,none.tsv?download=1}{\link} \\ \midrule
\multirow{2}{*}{NLI} 
& \textit{PAWS-X}  & \href{https://zenodo.org/records/13750485/files/evaluation_table_0shot_category-3_paws_acc,none.tsv?download=1}{\link} & \href{https://zenodo.org/records/13750495/files/evaluation_table_1shot_category-3_paws_acc,none.tsv?download=1}{\link} & \href{https://zenodo.org/records/13750497/files/evaluation_table_5shot_category-3_paws_acc,none.tsv?download=1}{\link} \\ 
& \textit{XNLI}  & \href{https://zenodo.org/records/13750485/files/evaluation_table_0shot_category-3_xnli_acc,none.tsv?download=1}{\link} & \href{https://zenodo.org/records/13750495/files/evaluation_table_1shot_category-3_xnli_acc,none.tsv?download=1}{\link} & \href{https://zenodo.org/records/13750497/files/evaluation_table_5shot_category-3_xnli_acc,none.tsv?download=1}{\link} \\ \midrule
Reading
& \textit{LAMBADA}  & \href{https://zenodo.org/records/13750485/files/evaluation_table_0shot_category-3_lambada_openai_mt_stablelm_acc,none.tsv?download=1}{\link} & \href{https://zenodo.org/records/13750495/files/evaluation_table_1shot_category-3_lambada_openai_mt_stablelm_acc,none.tsv?download=1}{\link} & \href{https://zenodo.org/records/13750497/files/evaluation_table_5shot_category-3_lambada_openai_mt_stablelm_acc,none.tsv?download=1}{\link} \\ 
Comprehension
& \textit{Belebele}  & \href{https://zenodo.org/records/13750485/files/evaluation_table_0shot_category-3_belebele_acc,none.tsv?download=1}{\link} & \href{https://zenodo.org/records/13750495/files/evaluation_table_1shot_category-3_belebele_acc,none.tsv?download=1}{\link} & \href{https://zenodo.org/records/13750497/files/evaluation_table_5shot_category-3_belebele_acc,none.tsv?download=1}{\link} \\ 
\bottomrule
\end{tabular}
}
\caption{Links to \textit{Category-3} models evaluation results for each task in zero-shot, one-shot, and few-shot setting.}
\label{tab:cat3-eval}
\end{table}
\begin{table}[h]
\centering
\resizebox{\linewidth}{!}{%
\begin{tabular}{l l c c c}
\toprule
\textbf{Type} & \textbf{Task} & \textbf{0-shot} & \textbf{1-shot} & \textbf{few-shot} \\
\midrule
\multirow{5}{*}{Q\&A} 
& \textit{ARC}  & \href{https://zenodo.org/records/13750485/files/evaluation_table_0shot_all_arc_acc,none.tsv?download=1}{\link} & \href{https://zenodo.org/records/13750495/files/evaluation_table_1shot_all_arc_acc,none.tsv?download=1}{\link} & \href{https://zenodo.org/records/13750497/files/evaluation_table_5shot_all_arc_acc,none.tsv?download=1}{\link} \\ 
& \textit{MGSM (Direct)}  & \href{https://zenodo.org/records/13750485/files/evaluation_table_0shot_all_mgsm_direct_exact_match,flexible-extract.tsv?download=1}{\link} & \href{https://zenodo.org/records/13750495/files/evaluation_table_1shot_all_mgsm_direct_exact_match,flexible-extract.tsv?download=1}{\link} & \href{https://zenodo.org/records/13750497/files/evaluation_table_5shot_all_mgsm_direct_exact_match,flexible-extract.tsv?download=1}{\link} \\ 
& \textit{MGSM (Native CoT)}  & \href{https://zenodo.org/records/13750485/files/evaluation_table_0shot_all_mgsm_native_cot_exact_match,flexible-extract.tsv?download=1}{\link} & \href{https://zenodo.org/records/13750495/files/evaluation_table_1shot_all_mgsm_native_cot_exact_match,flexible-extract.tsv?download=1}{\link} & \href{https://zenodo.org/records/13750497/files/evaluation_table_5shot_all_mgsm_native_cot_exact_match,flexible-extract.tsv?download=1}{\link} \\ 
& \textit{TruthfulQA}  & \href{https://zenodo.org/records/13750485/files/evaluation_table_0shot_all_truthfulqa_mc1_acc,none.tsv?download=1}{\link} & \href{https://zenodo.org/records/13750495/files/evaluation_table_1shot_all_truthfulqa_mc1_acc,none.tsv?download=1}{\link} & \href{https://zenodo.org/records/13750497/files/evaluation_table_5shot_all_truthfulqa_mc1_acc,none.tsv?download=1}{\link} \\ 
& \textit{MMLU}  & \href{https://zenodo.org/records/13750485/files/evaluation_table_0shot_all_m_mmlu_acc,none.tsv?download=1}{\link} & \href{https://zenodo.org/records/13750495/files/evaluation_table_1shot_all_m_mmlu_acc,none.tsv?download=1}{\link} & \href{https://zenodo.org/records/13750497/files/evaluation_table_5shot_all_m_mmlu_acc,none.tsv?download=1}{\link} \\ \midrule
\multirow{4}{*}{Reasoning} 
& \textit{HellaSwag}  & \href{https://zenodo.org/records/13750485/files/evaluation_table_0shot_all_hellaswag_acc,none.tsv?download=1}{\link} & \href{https://zenodo.org/records/13750495/files/evaluation_table_1shot_all_hellaswag_acc,none.tsv?download=1}{\link} & \href{https://zenodo.org/records/13750497/files/evaluation_table_5shot_all_hellaswag_acc,none.tsv?download=1}{\link} \\ 
& \textit{XCOPA}  & \href{https://zenodo.org/records/13750485/files/evaluation_table_0shot_all_xcopa_acc,none.tsv?download=1}{\link} & \href{https://zenodo.org/records/13750495/files/evaluation_table_1shot_all_xcopa_acc,none.tsv?download=1}{\link} & \href{https://zenodo.org/records/13750497/files/evaluation_table_5shot_all_xcopa_acc,none.tsv?download=1}{\link} \\ 
& \textit{XStoryCloze}  & \href{https://zenodo.org/records/13750485/files/evaluation_table_0shot_all_xstorycloze_acc,none.tsv?download=1}{\link} & \href{https://zenodo.org/records/13750495/files/evaluation_table_1shot_all_xstorycloze_acc,none.tsv?download=1}{\link} & \href{https://zenodo.org/records/13750497/files/evaluation_table_5shot_all_xstorycloze_acc,none.tsv?download=1}{\link} \\ 
& \textit{XWinograd}  & \href{https://zenodo.org/records/13750485/files/evaluation_table_0shot_all_xwinograd_acc,none.tsv?download=1}{\link} & \href{https://zenodo.org/records/13750495/files/evaluation_table_1shot_all_xwinograd_acc,none.tsv?download=1}{\link} & \href{https://zenodo.org/records/13750497/files/evaluation_table_5shot_all_xwinograd_acc,none.tsv?download=1}{\link} \\ \midrule
\multirow{2}{*}{NLI} 
& \textit{PAWS-X}  & \href{https://zenodo.org/records/13750485/files/evaluation_table_0shot_all_paws_acc,none.tsv?download=1}{\link} & \href{https://zenodo.org/records/13750495/files/evaluation_table_1shot_all_paws_acc,none.tsv?download=1}{\link} & \href{https://zenodo.org/records/13750497/files/evaluation_table_5shot_all_paws_acc,none.tsv?download=1}{\link} \\ 
& \textit{XNLI}  & \href{https://zenodo.org/records/13750485/files/evaluation_table_0shot_all_xnli_acc,none.tsv?download=1}{\link} & \href{https://zenodo.org/records/13750495/files/evaluation_table_1shot_all_xnli_acc,none.tsv?download=1}{\link} & \href{https://zenodo.org/records/13750497/files/evaluation_table_5shot_all_xnli_acc,none.tsv?download=1}{\link} \\ \midrule
Reading 
& \textit{LAMBADA}  & \href{https://zenodo.org/records/13750485/files/evaluation_table_0shot_all_lambada_openai_mt_stablelm_acc,none.tsv?download=1}{\link} & \href{https://zenodo.org/records/13750495/files/evaluation_table_1shot_all_lambada_openai_mt_stablelm_acc,none.tsv?download=1}{\link} & \href{https://zenodo.org/records/13750497/files/evaluation_table_5shot_all_lambada_openai_mt_stablelm_acc,none.tsv?download=1}{\link} \\ 
Comprehension
& \textit{Belebele}  & \href{https://zenodo.org/records/13750485/files/evaluation_table_0shot_all_belebele_acc,none.tsv?download=1}{\link} & \href{https://zenodo.org/records/13750495/files/evaluation_table_1shot_all_belebele_acc,none.tsv?download=1}{\link} & \href{https://zenodo.org/records/13750497/files/evaluation_table_5shot_all_belebele_acc,none.tsv?download=1}{\link} \\ 
\bottomrule
\end{tabular}
}
\caption{Links to combined (all model categories) evaluation results for each task in zero-shot, one-shot, and few-shot setting.}
\label{tab:all-eval}
\end{table}

\subsubsection{Missing Results}
\label{subsubsec:missing-results}
We acknowledge certain limitations in our experimental setup, particularly where some tables lack results for specific models under certain configurations.\footnote{All the files regarding the missing combinations can be found here: \href{https://zenodo.org/records/13763520}{zenodo.org/records/13763520}} We have identified two primary factors contributing to this: 1) Some models (\eg Glot500m, XLM-R) exhibit limitations in their tokenization logic, which causes them to fail on larger sequences or languages that they cannot tokenize properly; and 2) Certain model architectures (\eg mT5, mBART, uMT5) are not supported by task implementations such as \textit{Belebele}. We ensure that the missing results do not negatively impact the comparisons by excluding cases where results are unavailable. Additionally, we take care to prevent these missing results from skewing the visualizations in our favor by accurately representing them as valid gaps.\\

\subsection{Extended Results Analysis}
\label{subsec:extended-analysis}

In addition to the analysis of \approach's overall comparative performance on all tasks and languages combined (see \autoref{subsec:res-analysis}). We also analyze its comparative performance on each type of task. Figures \ref{fig:qa-perf}--\ref{fig:rc-perf} show the performance of \approach on each of the four task types, with the plots on the left representing comparisons based on the Wilcoxon signed-rank test and the plots on the right compare average performance across all languages. We also provide individual task-based analysis, where Figures \ref{fig:arc-perf}--\ref{fig:xwinograd-perf} show \approach's performance on each of the 13 different tasks. In the following subsections, we use these plots to discuss the performance of our system on each of these task types in detail.

\subsubsection{Q\&A}

\begin{figure*}[h]
  \centering
  \begin{minipage}{0.49\linewidth}
    \centering
    \includegraphics[width=\linewidth]{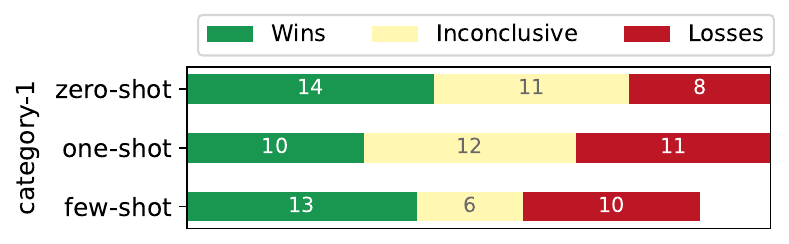}\\
    \includegraphics[width=\linewidth]{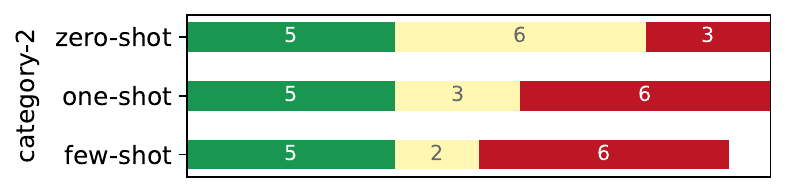}\\
    \includegraphics[width=\linewidth]{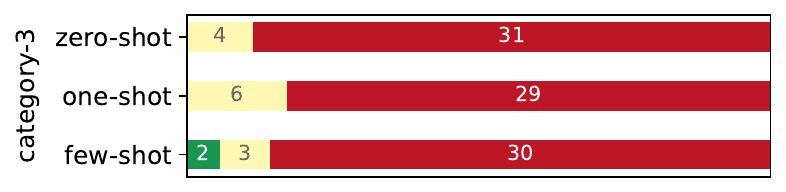}
  \end{minipage}
  \hfill
  \begin{minipage}{0.49\linewidth}
    \centering
    \includegraphics[width=\linewidth]{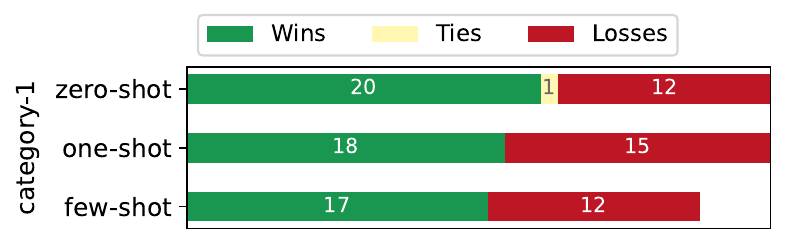}\\
    \includegraphics[width=\linewidth]{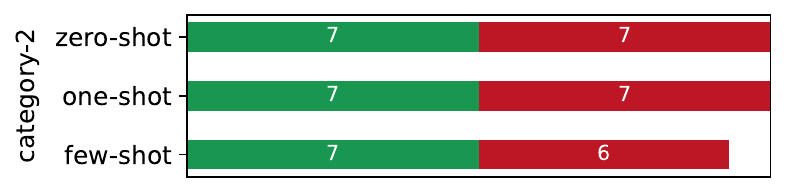}\\
    \includegraphics[width=\linewidth]{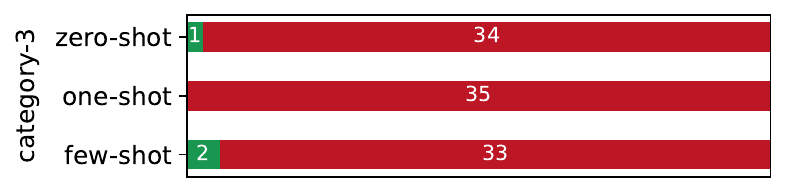}
  \end{minipage}
  \caption{Performance comparison for task type: \textbf{Q\&A} }
  \label{fig:qa-perf}
\end{figure*}

\begin{figure*}[h]
  \centering
  \begin{minipage}{0.49\linewidth}
    \centering
    \includegraphics[width=\linewidth]{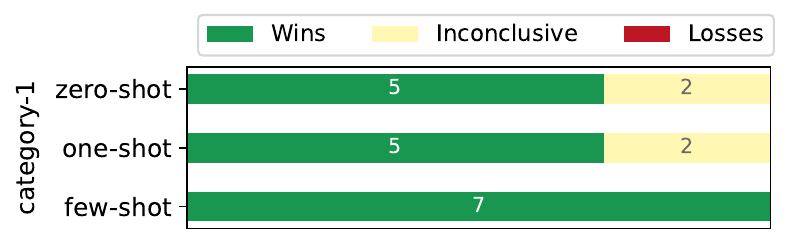}\\
    \includegraphics[width=\linewidth]{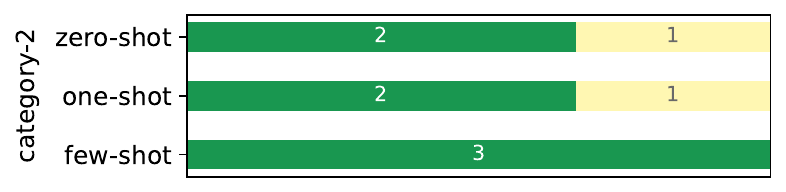}\\
    \includegraphics[width=\linewidth]{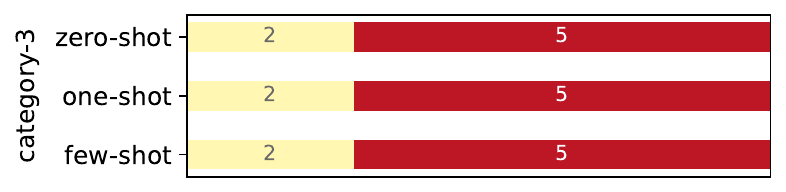}
  \end{minipage}
  \hfill
  \begin{minipage}{0.49\linewidth}
    \centering
    \includegraphics[width=\linewidth]{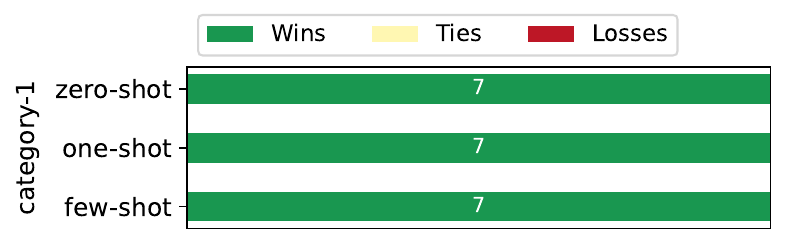}\\
    \includegraphics[width=\linewidth]{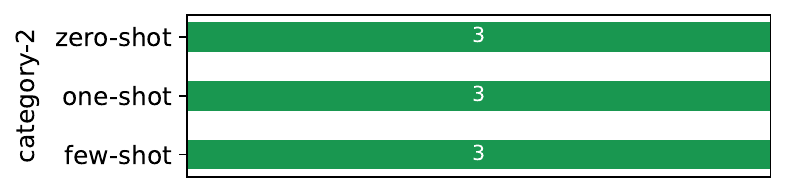}\\
    \includegraphics[width=\linewidth]{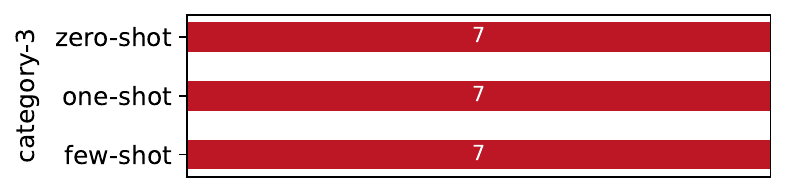}
  \end{minipage}
  \caption{Performance comparison for task: \textbf{\textit{ARC}} }
  \label{fig:arc-perf}
\end{figure*}

\begin{figure*}[h]
  \centering
  \begin{minipage}{0.49\linewidth}
    \centering
    \includegraphics[width=\linewidth]{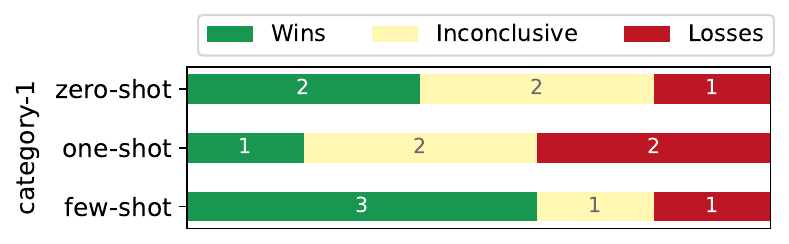}\\
    \includegraphics[width=\linewidth]{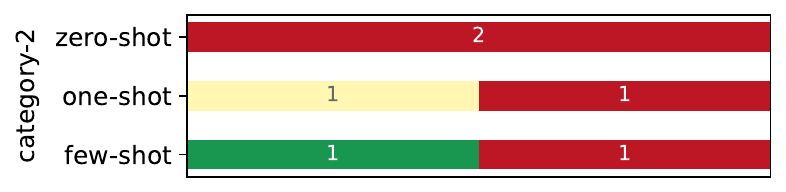}\\
    \includegraphics[width=\linewidth]{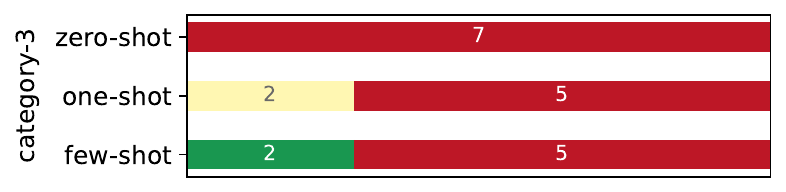}
  \end{minipage}
  \hfill
  \begin{minipage}{0.49\linewidth}
    \centering
    \includegraphics[width=\linewidth]{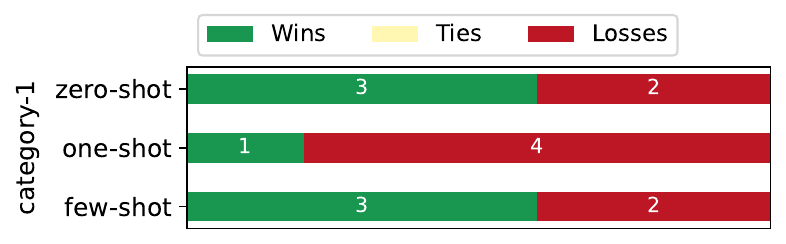}\\
    \includegraphics[width=\linewidth]{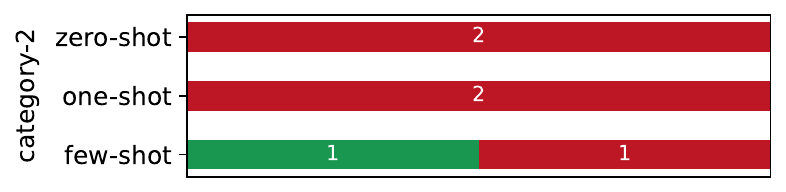}\\
    \includegraphics[width=\linewidth]{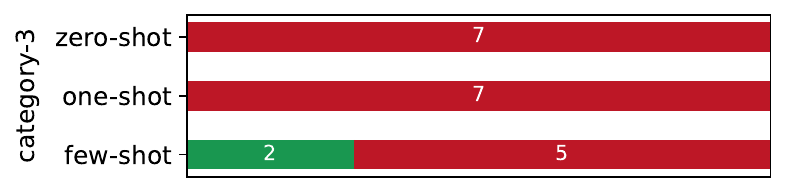}
  \end{minipage}
  \caption{Performance comparison for task: \textbf{\textit{MMLU}} }
  \label{fig:m_mmlu-perf}
\end{figure*}

\begin{figure*}[h]
  \centering
  \begin{minipage}{0.49\linewidth}
    \centering
    \includegraphics[width=\linewidth]{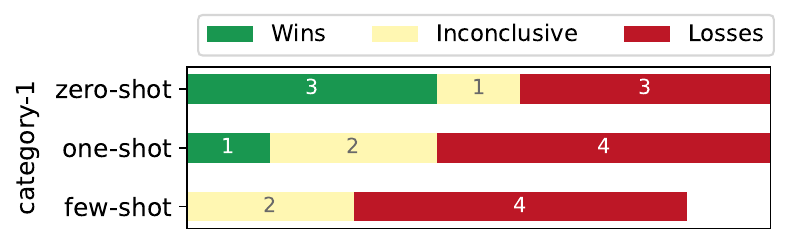}\\
    \includegraphics[width=\linewidth]{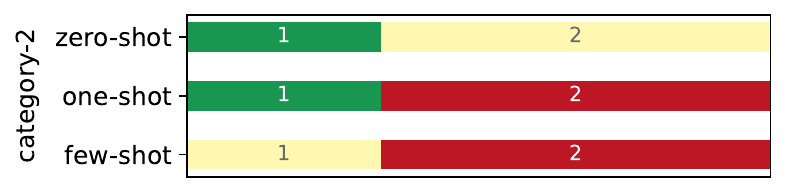}\\
    \includegraphics[width=\linewidth]{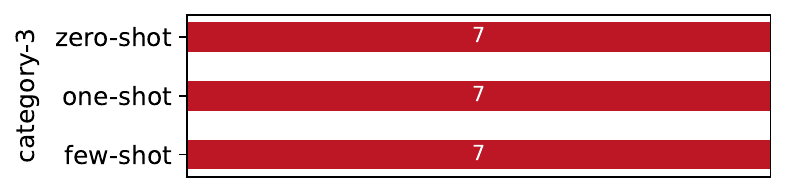}
  \end{minipage}
  \hfill
  \begin{minipage}{0.49\linewidth}
    \centering
    \includegraphics[width=\linewidth]{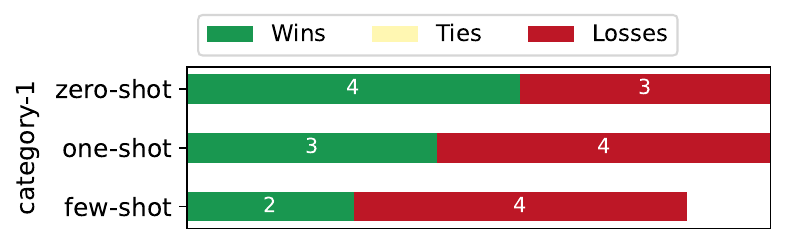}\\
    \includegraphics[width=\linewidth]{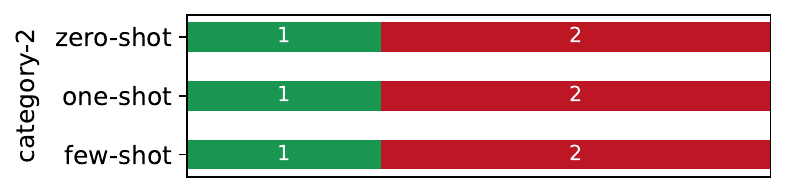}\\
    \includegraphics[width=\linewidth]{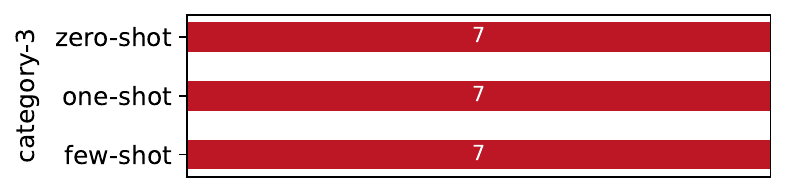}
  \end{minipage}
  \caption{Performance comparison for task: \textbf{\textit{MGSM (Direct)}} }
  \label{fig:mgsm_direct-perf}
\end{figure*}

\begin{figure*}[h]
  \centering
  \begin{minipage}{0.49\linewidth}
    \centering
    \includegraphics[width=\linewidth]{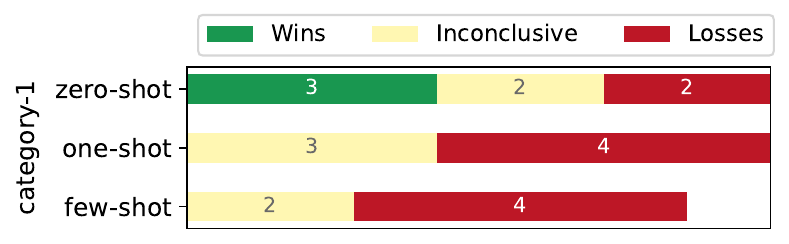}\\
    \includegraphics[width=\linewidth]{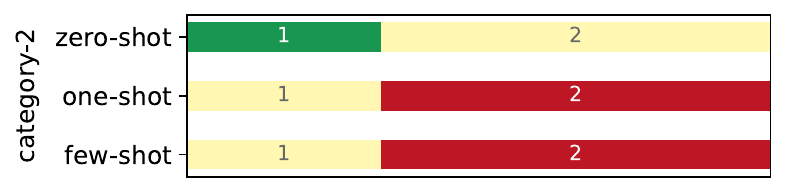}\\
    \includegraphics[width=\linewidth]{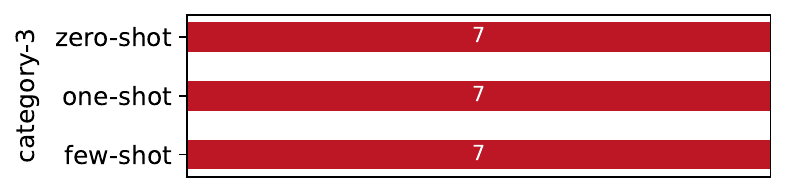}
  \end{minipage}
  \hfill
  \begin{minipage}{0.49\linewidth}
    \centering
    \includegraphics[width=\linewidth]{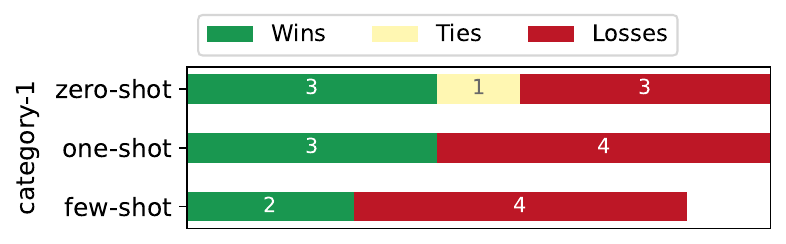}\\
    \includegraphics[width=\linewidth]{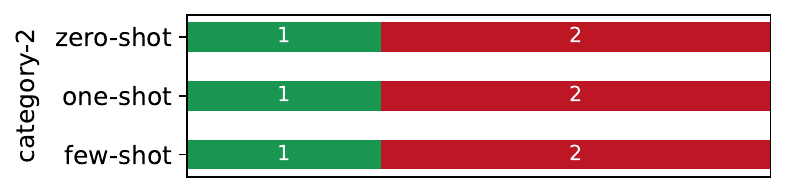}\\
    \includegraphics[width=\linewidth]{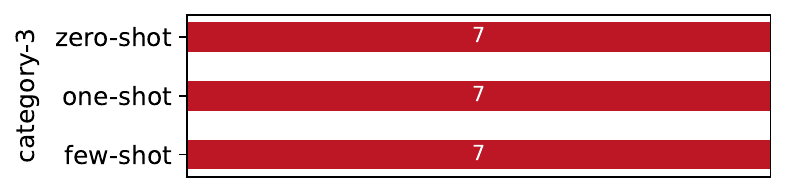}
  \end{minipage}
  \caption{Performance comparison for task: \textbf{\textit{MGSM (Native CoT)}} }
  \label{fig:mgsm_native_cot-perf}
\end{figure*}

\begin{figure*}[h]
  \centering
  \begin{minipage}{0.49\linewidth}
    \centering
    \includegraphics[width=\linewidth]{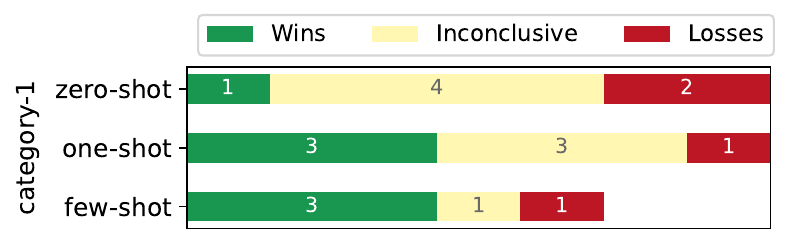}\\
    \includegraphics[width=\linewidth]{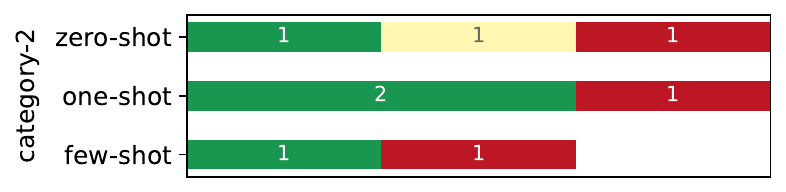}\\
    \includegraphics[width=\linewidth]{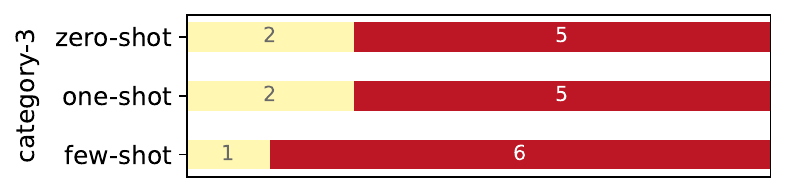}
  \end{minipage}
  \hfill
  \begin{minipage}{0.49\linewidth}
    \centering
    \includegraphics[width=\linewidth]{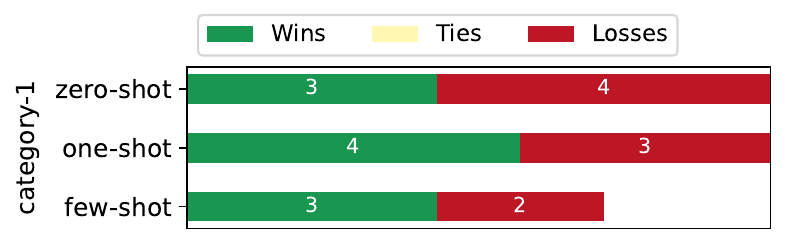}\\
    \includegraphics[width=\linewidth]{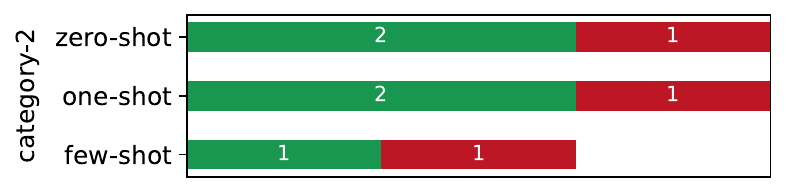}\\
    \includegraphics[width=\linewidth]{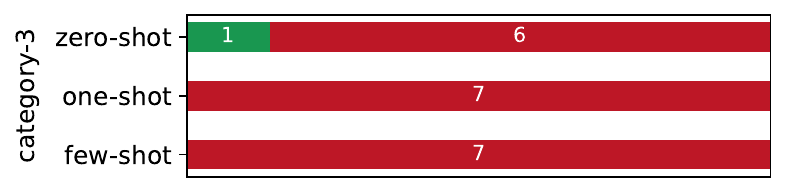}
  \end{minipage}
  \caption{Performance comparison for task: \textbf{\textit{TruthfulQA}} }
  \label{fig:truthfulqa-perf}
\end{figure*}

As observed from \autoref{fig:qa-perf}, \approach has balanced performance on both significance and average comparison metrics against \textit{Category-1} and \textit{Category-2} models. However, in comparison to the \textit{Category-3} models, it performs poorly. Also, we notice that the performance of our model decreases in the one-shot and few-shot settings. Looking closer at the individual tasks, we find that on the \textit{ARC} (see \autoref{fig:arc-perf}), it demonstrates strong performance against \textit{Category-1} and \textit{Category-2}, while exhibiting significantly weaker performance against \textit{Category-3}. In contrast, on \textit{MGSM} (see \autoref{fig:mgsm_direct-perf} and \autoref{fig:mgsm_native_cot-perf}), it performs poorly against \textit{Category-1} and \textit{Category-2}, and is comprehensively outperformed in \textit{Category-3}. For MMLU (see \autoref{fig:m_mmlu-perf}), it shows balanced performance in \textit{Category-1} but struggles with weaker results in \textit{Category-2} and \textit{Category-3}. Lastly, on TruthfulQA (see \autoref{fig:truthfulqa-perf}), it maintains balanced performance for \textit{Category-1} and \textit{Category-2}, but shows a noticeable weakness in \textit{Category-3}.

\subsubsection{Reasoning}

\begin{figure*}[h]
  \centering
  \begin{minipage}{0.49\linewidth}
    \centering
    \includegraphics[width=\linewidth]{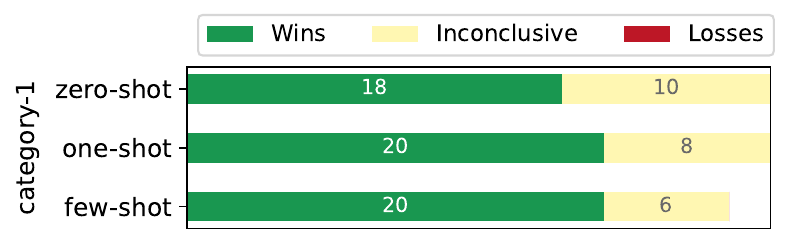}\\
    \includegraphics[width=\linewidth]{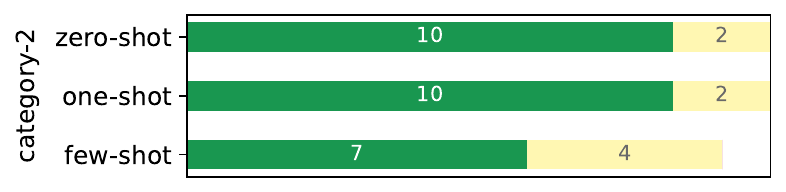}\\
    \includegraphics[width=\linewidth]{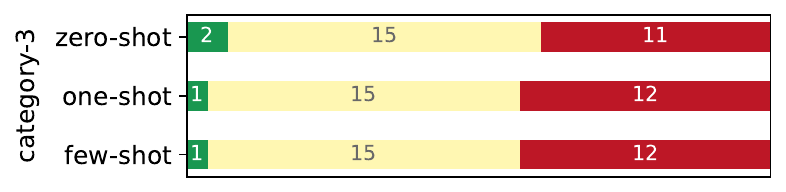}
  \end{minipage}
  \hfill
  \begin{minipage}{0.49\linewidth}
    \centering
    \includegraphics[width=\linewidth]{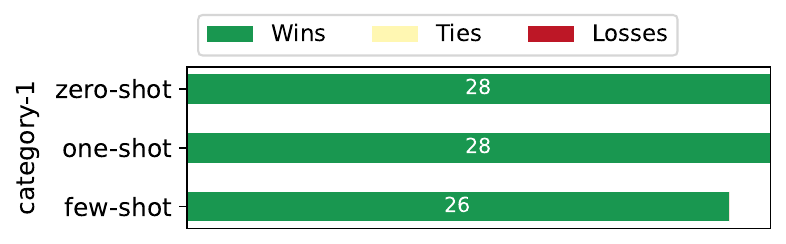}\\
    \includegraphics[width=\linewidth]{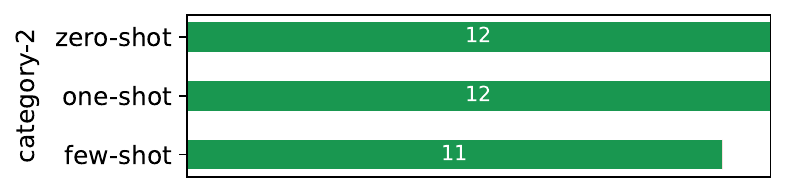}\\
    \includegraphics[width=\linewidth]{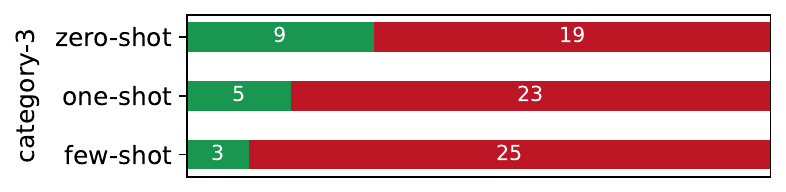}
  \end{minipage}
  \caption{Performance comparison for task type: \textbf{Reasoning} }
  \label{fig:reasoning-perf}
\end{figure*}

\begin{figure*}[h]
  \centering
  \begin{minipage}{0.49\linewidth}
    \centering
    \includegraphics[width=\linewidth]{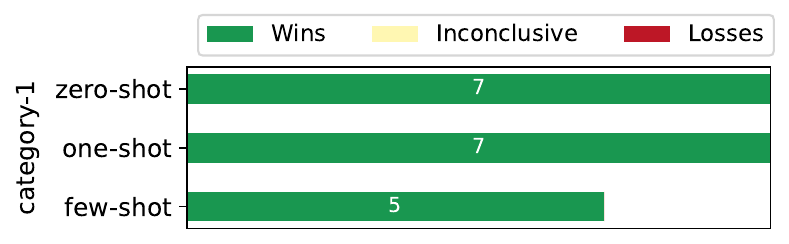}\\
    \includegraphics[width=\linewidth]{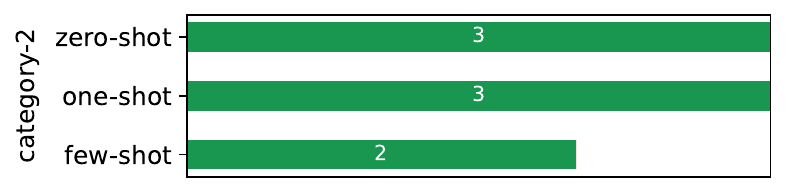}\\
    \includegraphics[width=\linewidth]{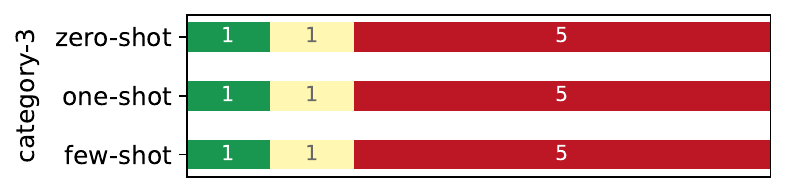}
  \end{minipage}
  \hfill
  \begin{minipage}{0.49\linewidth}
    \centering
    \includegraphics[width=\linewidth]{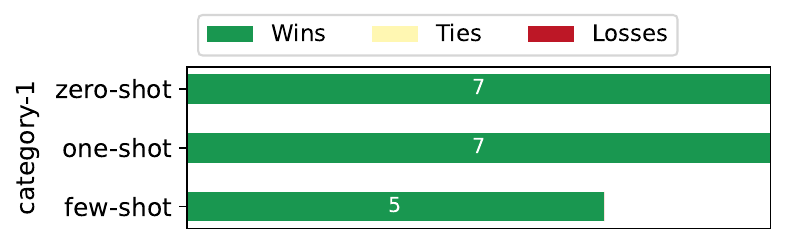}\\
    \includegraphics[width=\linewidth]{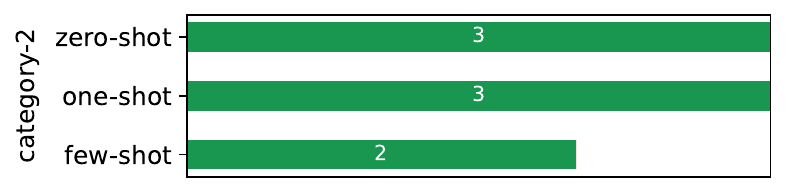}\\
    \includegraphics[width=\linewidth]{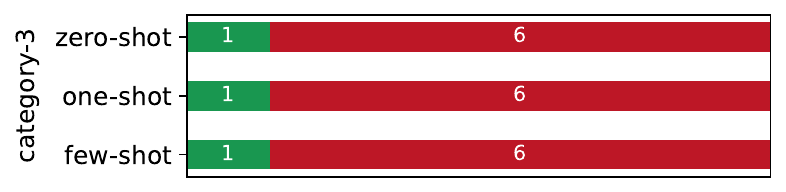}
  \end{minipage}
  \caption{Performance comparison for task: \textbf{\textit{HellaSwag}} }
  \label{fig:hellaswag-perf}
\end{figure*}

\begin{figure*}[h]
  \centering
  \begin{minipage}{0.49\linewidth}
    \centering
    \includegraphics[width=\linewidth]{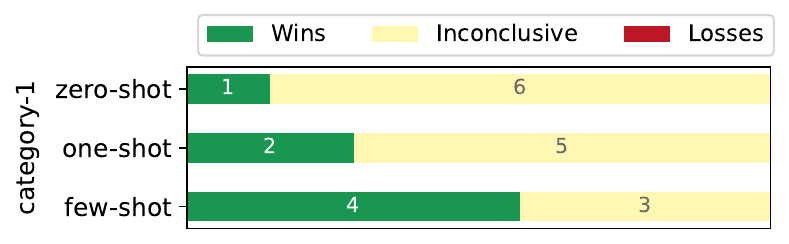}\\
    \includegraphics[width=\linewidth]{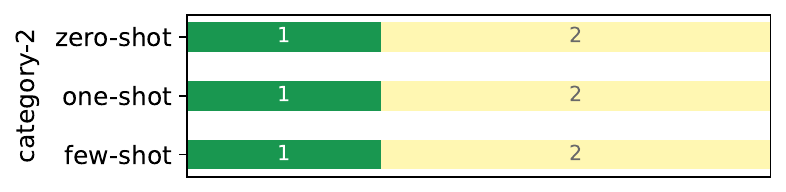}\\
    \includegraphics[width=\linewidth]{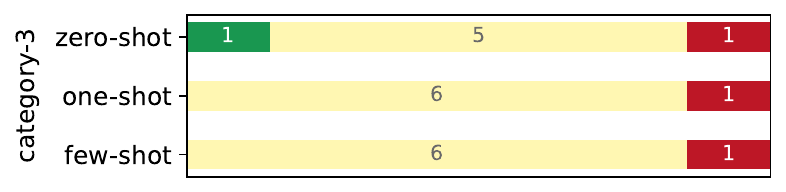}
  \end{minipage}
  \hfill
  \begin{minipage}{0.49\linewidth}
    \centering
    \includegraphics[width=\linewidth]{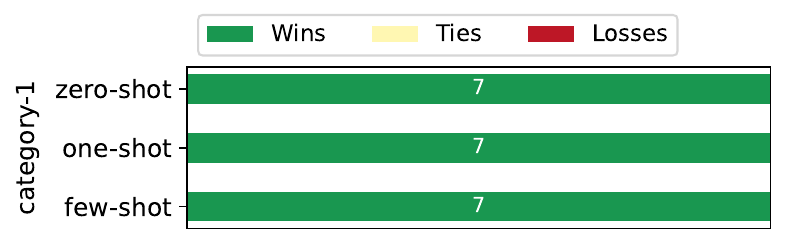}\\
    \includegraphics[width=\linewidth]{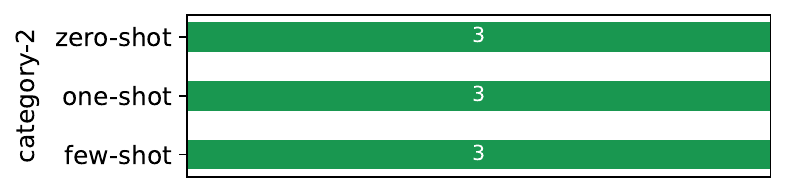}\\
    \includegraphics[width=\linewidth]{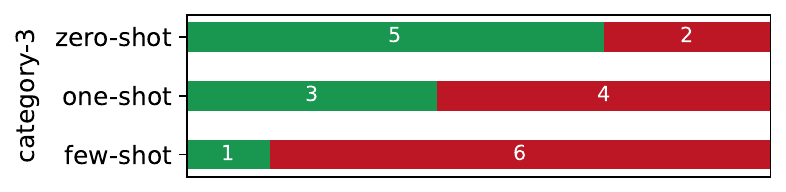}
  \end{minipage}
  \caption{Performance comparison for task: \textbf{\textit{XCOPA}} }
  \label{fig:xcopa-perf}
\end{figure*}

\begin{figure*}[h]
  \centering
  \begin{minipage}{0.49\linewidth}
    \centering
    \includegraphics[width=\linewidth]{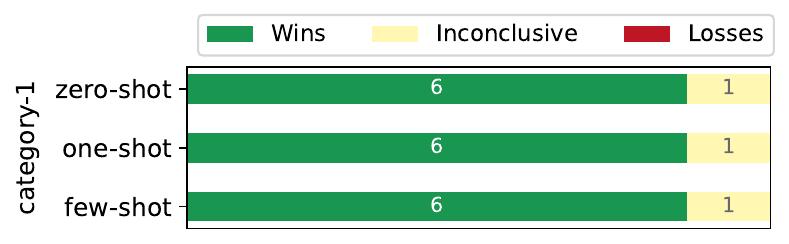}\\
    \includegraphics[width=\linewidth]{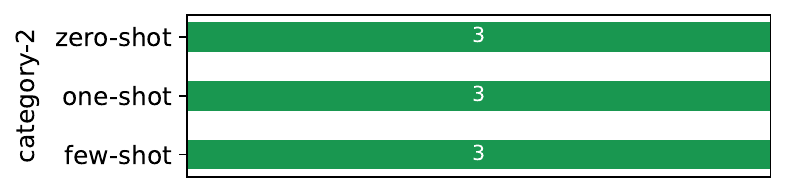}\\
    \includegraphics[width=\linewidth]{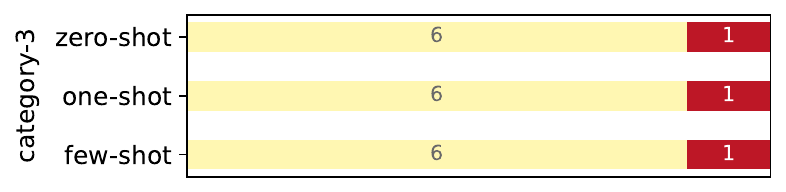}
  \end{minipage}
  \hfill
  \begin{minipage}{0.49\linewidth}
    \centering
    \includegraphics[width=\linewidth]{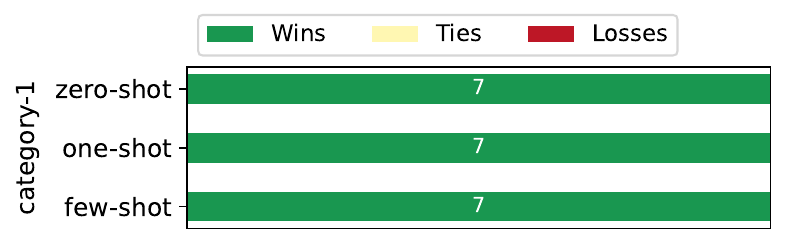}\\
    \includegraphics[width=\linewidth]{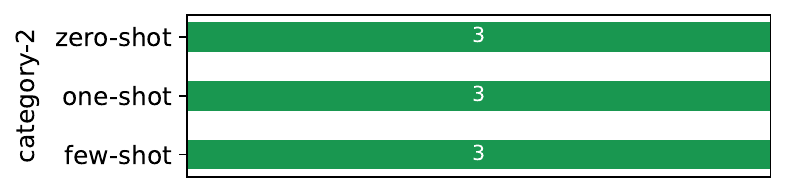}\\
    \includegraphics[width=\linewidth]{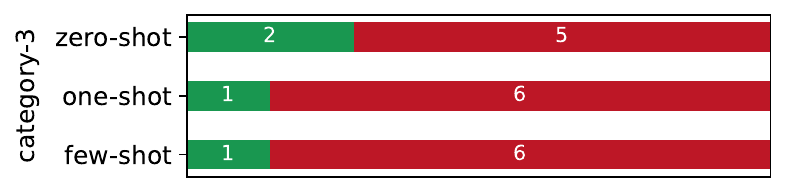}
  \end{minipage}
  \caption{Performance comparison for task: \textbf{\textit{XStorycloze}} }
  \label{fig:xstorycloze-perf}
\end{figure*}

\begin{figure*}[h]
  \centering
  \begin{minipage}{0.49\linewidth}
    \centering
    \includegraphics[width=\linewidth]{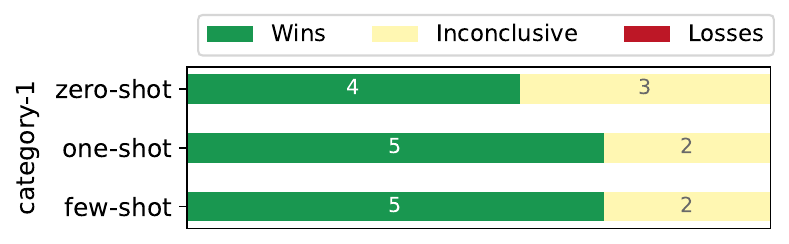}\\
    \includegraphics[width=\linewidth]{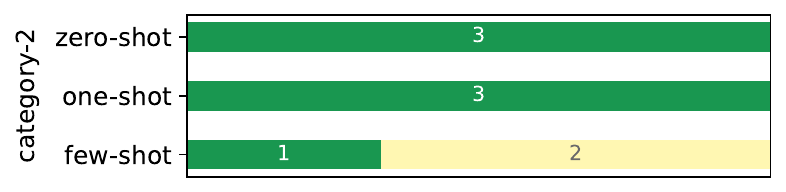}\\
    \includegraphics[width=\linewidth]{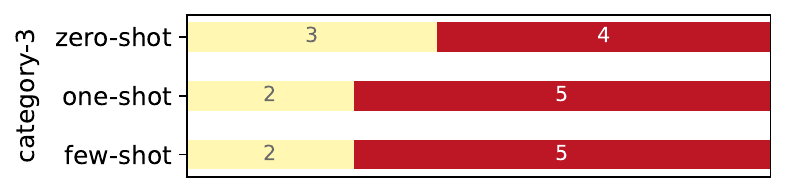}
  \end{minipage}
  \hfill
  \begin{minipage}{0.49\linewidth}
    \centering
    \includegraphics[width=\linewidth]{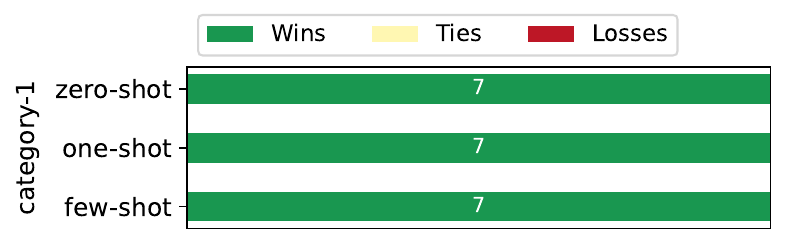}\\
    \includegraphics[width=\linewidth]{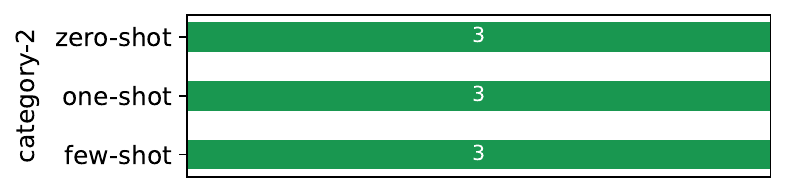}\\
    \includegraphics[width=\linewidth]{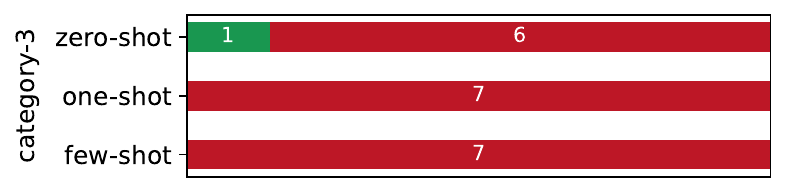}
  \end{minipage}
  \caption{Performance comparison for task: \textbf{\textit{XWinograd}} }
  \label{fig:xwinograd-perf}
\end{figure*}

In \autoref{fig:reasoning-perf}, we observe that \approach outperforms \textit{Category-1} and \textit{Category-2} models comprehensively. However, it shows weak performance against \textit{Category-3}. Looking further at the individual tasks, for \textit{HellaSwag} (see \autoref{fig:hellaswag-perf}), it demonstrates good overall performance on \textit{Category-1} and \textit{Category-2}, but performs poorly on \textit{Category-3}. A similar pattern is observed for \textit{XWinograd} (see \autoref{fig:xwinograd-perf}) as well. On XCOPA (see \autoref{fig:xcopa-perf}), it shows strong results for \textit{Category-1} and \textit{Category-2}, with mostly inconclusive significance results on \textit{Category-3}, although it achieves better average performance in the zero-shot setting against \textit{Category-3}. Lastly, for XStoryCloze (see \autoref{fig:xstorycloze-perf}), it performs well on \textit{Category-1} and \textit{Category-2}, but shows mostly inconclusive significance results and consistently loses in average performance on \textit{Category-3}.

\subsubsection{NLI}

\begin{figure*}[h]
  \centering
  \begin{minipage}{0.49\linewidth}
    \centering
    \includegraphics[width=\linewidth]{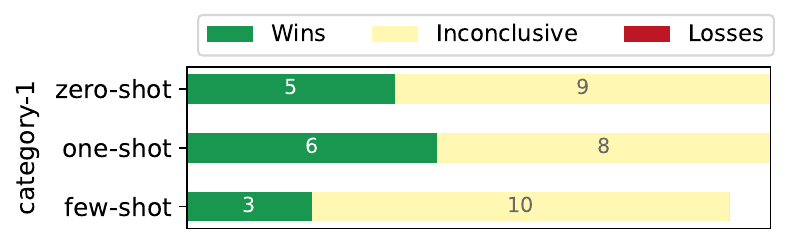}\\
    \includegraphics[width=\linewidth]{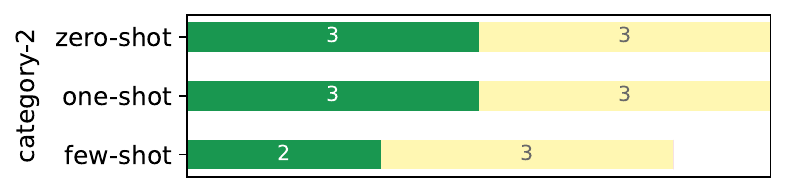}\\
    \includegraphics[width=\linewidth]{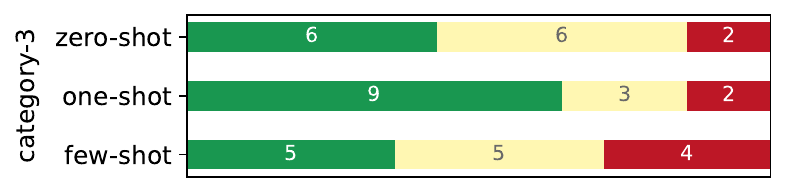}
  \end{minipage}
  \hfill
  \begin{minipage}{0.49\linewidth}
    \centering
    \includegraphics[width=\linewidth]{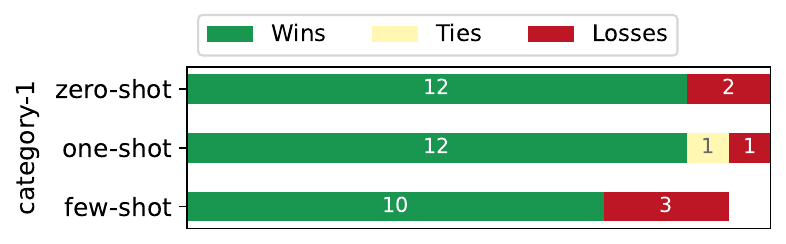}\\
    \includegraphics[width=\linewidth]{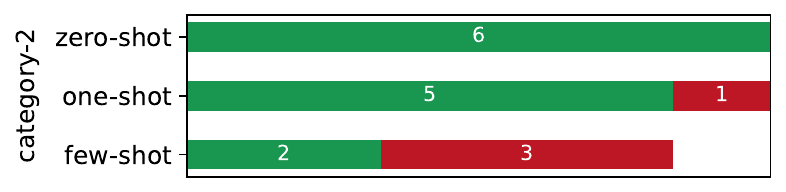}\\
    \includegraphics[width=\linewidth]{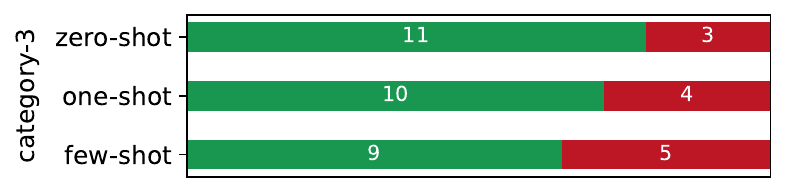}
  \end{minipage}
  \caption{Performance comparison for task type: \textbf{NLI} }
  \label{fig:nli-perf}
\end{figure*}

\begin{figure*}[h]
  \centering
  \begin{minipage}{0.49\linewidth}
    \centering
    \includegraphics[width=\linewidth]{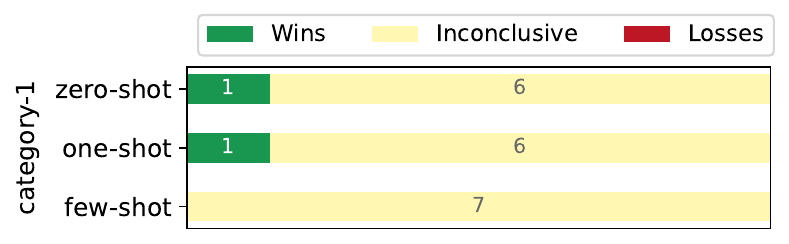}\\
    \includegraphics[width=\linewidth]{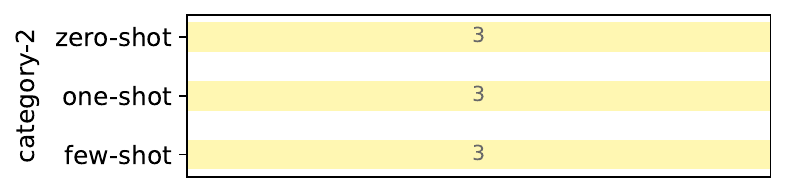}\\
    \includegraphics[width=\linewidth]{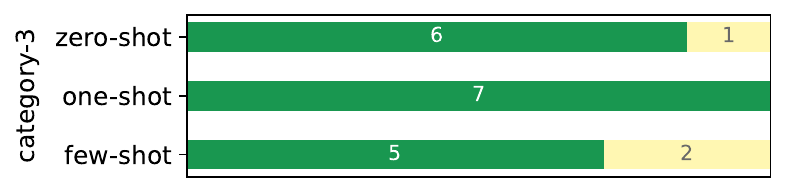}
  \end{minipage}
  \hfill
  \begin{minipage}{0.49\linewidth}
    \centering
    \includegraphics[width=\linewidth]{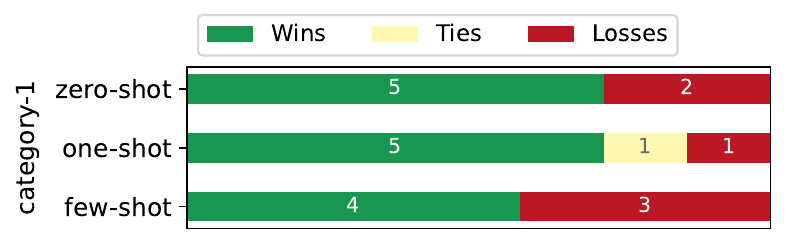}\\
    \includegraphics[width=\linewidth]{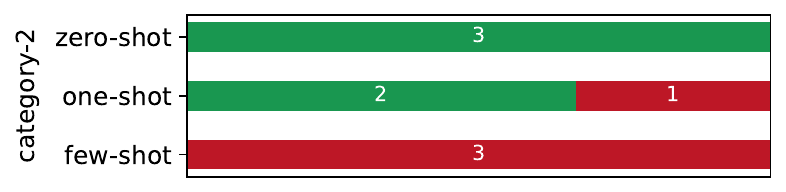}\\
    \includegraphics[width=\linewidth]{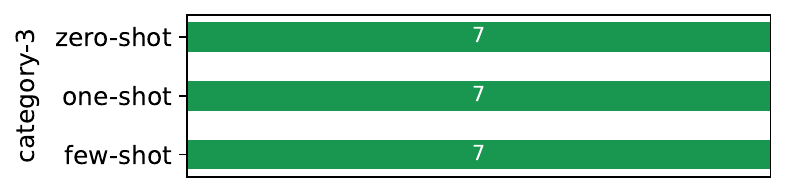}
  \end{minipage}
  \caption{Performance comparison for task: \textbf{\textit{PAWS-X}} }
  \label{fig:paws-perf}
\end{figure*}

\begin{figure*}[h]
  \centering
  \begin{minipage}{0.49\linewidth}
    \centering
    \includegraphics[width=\linewidth]{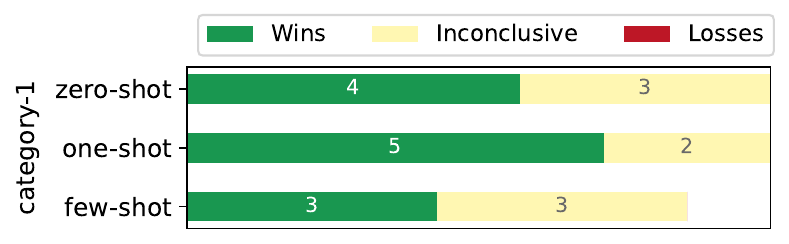}\\
    \includegraphics[width=\linewidth]{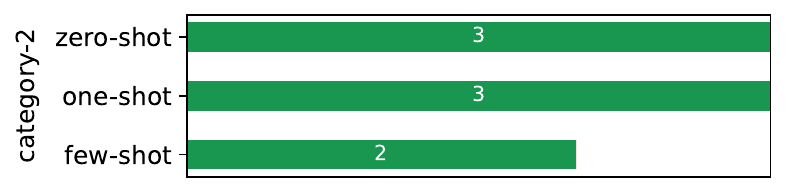}\\
    \includegraphics[width=\linewidth]{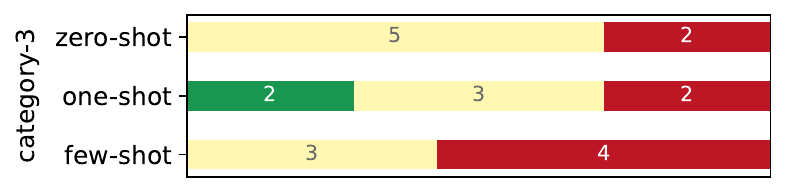}
  \end{minipage}
  \hfill
  \begin{minipage}{0.49\linewidth}
    \centering
    \includegraphics[width=\linewidth]{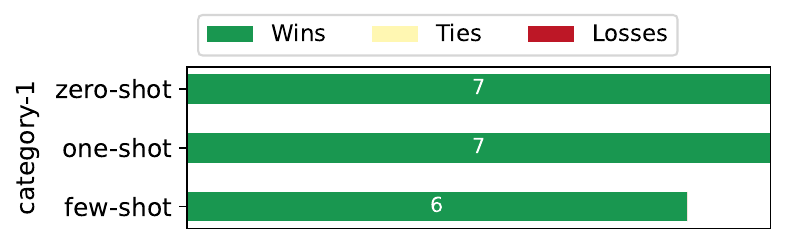}\\
    \includegraphics[width=\linewidth]{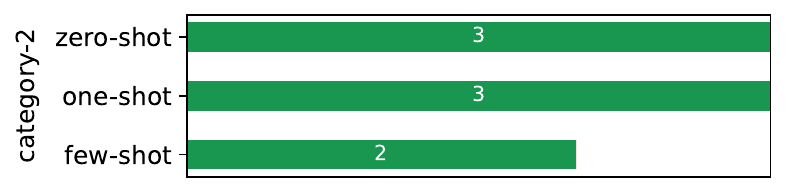}\\
    \includegraphics[width=\linewidth]{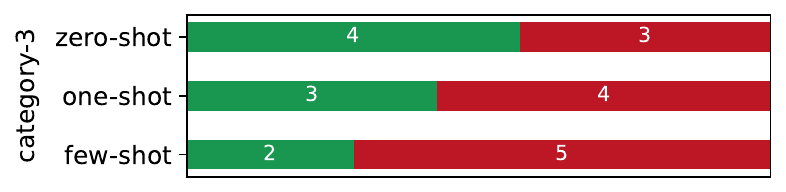}
  \end{minipage}
  \caption{Performance comparison for task: \textbf{\textit{XNLI}} }
  \label{fig:xnli-perf}
\end{figure*}

Looking at the overall NLI results in \autoref{fig:nli-perf}, we notice that \approach performs pretty well across all categories. On the individual tasks, we observe that for \textit{Paws-X} (see \autoref{fig:paws-perf}), significance results show inconclusive performance on \textit{Category-1} and \textit{Category-2}, but surprisingly, it performs overwhelmingly well against \textit{Category-3}. In terms of average performance, the model achieves good results for both \textit{Category-1} and \textit{Category-2}, while delivering a clean sweep in favor of \approach against \textit{Category-3}. On \textit{XNLI} (see \autoref{fig:xnli-perf}), it demonstrates very strong performance for \textit{Category-1} and \textit{Category-2}, though results for \textit{Category-3} are mostly inconclusive. However, the average performance across all categories remains balanced.

\subsubsection{Reading Comprehension}
\label{subsubsec:rc-analysis}
\begin{figure*}[h]
  \centering
  \begin{minipage}{0.49\linewidth}
    \centering
    \includegraphics[width=\linewidth]{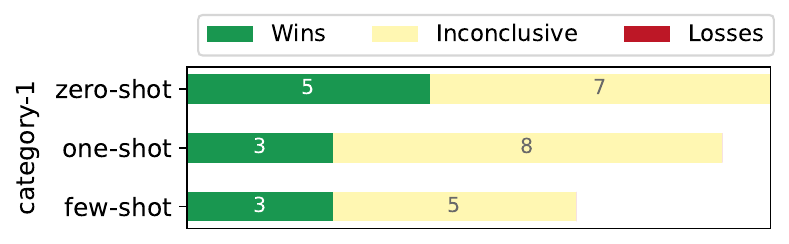}\\
    \includegraphics[width=\linewidth]{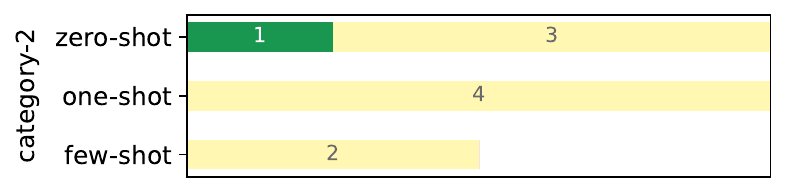}\\
    \includegraphics[width=\linewidth]{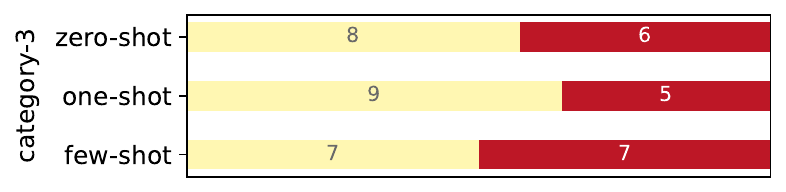}
  \end{minipage}
  \hfill
  \begin{minipage}{0.49\linewidth}
    \centering
    \includegraphics[width=\linewidth]{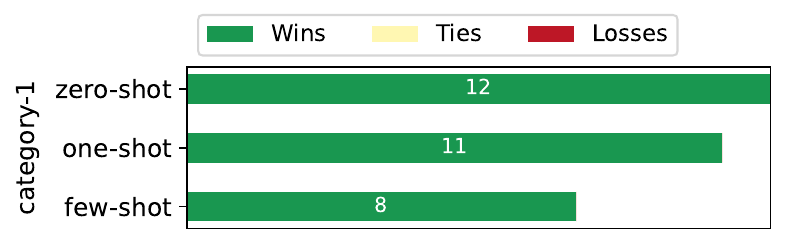}\\
    \includegraphics[width=\linewidth]{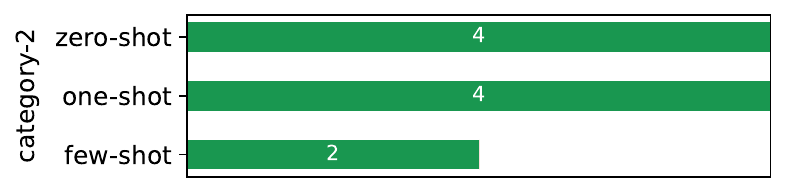}\\
    \includegraphics[width=\linewidth]{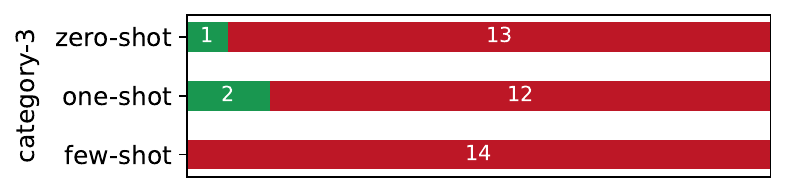}
  \end{minipage}
  \caption{Performance comparison for task type: \textbf{Reading Comprehension} }
  \label{fig:rc-perf}
\end{figure*}

\begin{figure*}[h]
  \centering
  \begin{minipage}{0.49\linewidth}
    \centering
    \includegraphics[width=\linewidth]{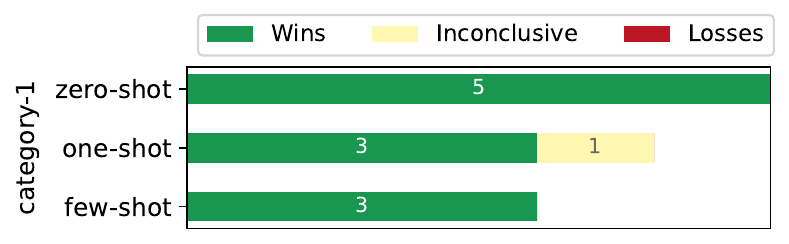}\\
    \includegraphics[width=\linewidth]{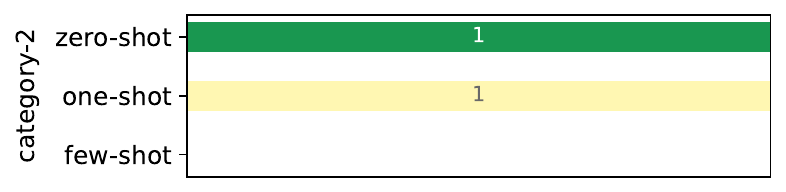}\\
    \includegraphics[width=\linewidth]{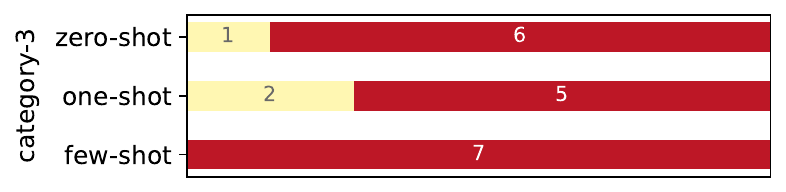}
  \end{minipage}
  \hfill
  \begin{minipage}{0.49\linewidth}
    \centering
    \includegraphics[width=\linewidth]{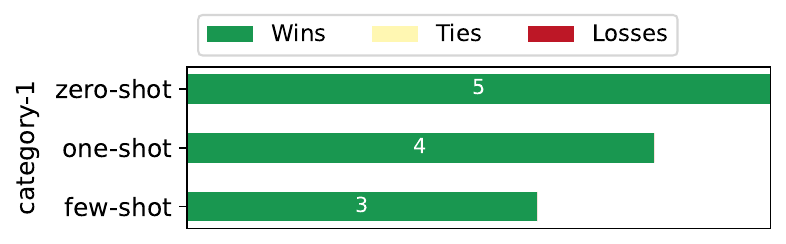}\\
    \includegraphics[width=\linewidth]{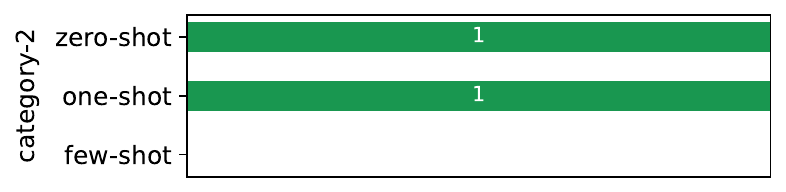}\\
    \includegraphics[width=\linewidth]{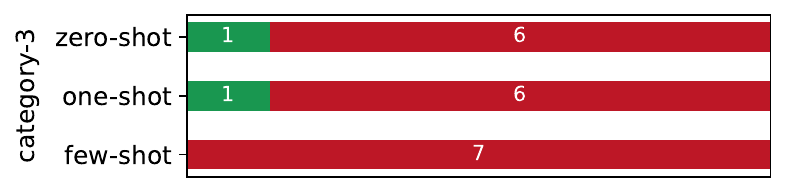}
  \end{minipage}
  \caption{Performance comparison for task: \textbf{\textit{Belebele}} }
  \label{fig:belebele-perf}
\end{figure*}

\begin{figure*}[h]
  \centering
  \begin{minipage}{0.49\linewidth}
    \centering
    \includegraphics[width=\linewidth]{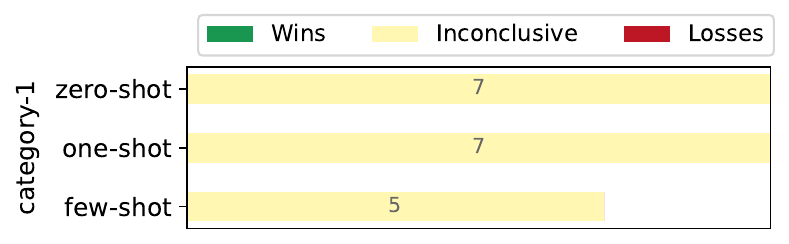}\\
    \includegraphics[width=\linewidth]{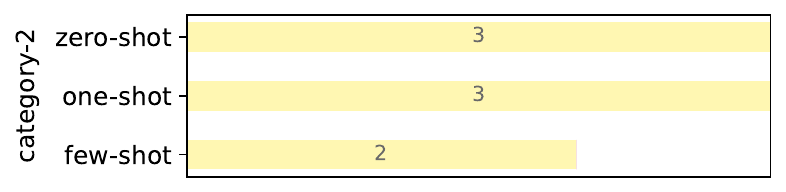}\\
    \includegraphics[width=\linewidth]{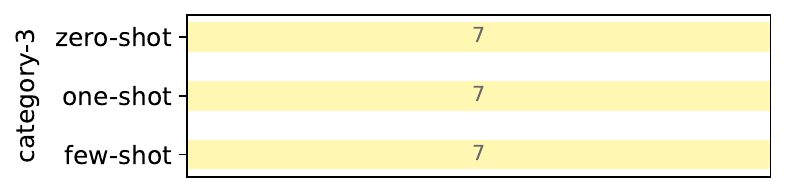}
  \end{minipage}
  \hfill
  \begin{minipage}{0.49\linewidth}
    \centering
    \includegraphics[width=\linewidth]{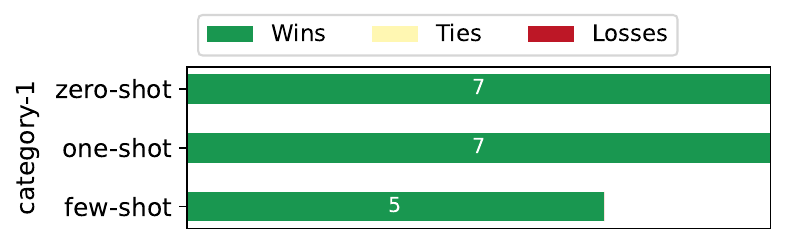}\\
    \includegraphics[width=\linewidth]{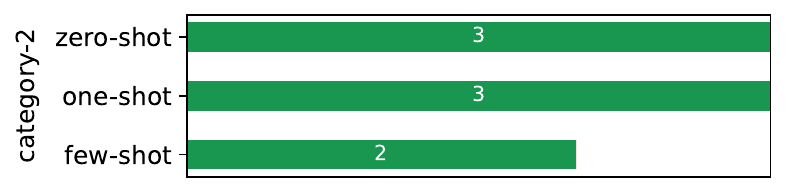}\\
    \includegraphics[width=\linewidth]{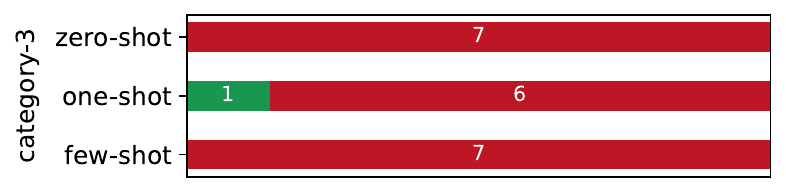}
  \end{minipage}
  \caption{Performance comparison for task: \textbf{\textit{LAMBADA}} }
  \label{fig:lambada-perf}
\end{figure*}

\autoref{fig:rc-perf} illustrates that there are many inconclusive significance comparisons for \textit{Category-1} and \textit{Category-2}, yet \approach completely outperforms in terms of average performance. However, similar to previous tasks, it shows weaker performance against \textit{Category-3}. For \textit{LAMBADA} (see \autoref{fig:lambada-perf}), significance comparisons across all categories yield only inconclusive results. Nevertheless, the average performance reveals that it dominates \textit{Category-1} and \textit{Category-2}, while being overwhelmingly outperformed in \textit{Category-3}. In \textit{Belebele} (see \autoref{fig:belebele-perf}), it demonstrates strong performance in \textit{Category-1}. However, due to the absence of support from two models for the task, \textit{Category-2} only allows comparison to a single model, against which our model performs well. In \textit{Category-3}, it again loses out to the others. \\

\noindent
To quickly summarize all the results across the various tasks, we observe that \approach generally performs well against \textit{Category-1} and \textit{Category-2}, while consistently showing weaker performance against \textit{Category-3}. In the Q\&A tasks, \approach maintains balanced performance in both significance and average comparison for \textit{Category-1} and \textit{Category-2}, but struggles in the one-shot and few-shot settings, particularly against \textit{Category-3}. For reasoning tasks, \approach demonstrates strong performance on \textit{Category-1} and \textit{Category-2}, with mixed results in \textit{Category-3}, where it achieves better average performance in zero-shot but falls short in other settings. In the NLI tasks, \approach performs strongly across all categories, with notable success in average performance against \textit{Category-3}, despite some inconclusive significance comparisons. Lastly, in reading comprehension tasks, while significance comparisons are often inconclusive for \textit{Category-1} and \textit{Category-2}, \approach still dominates in average performance but continues to struggle against \textit{Category-3}.

\subsection{Extended MoE Analysis}
\label{subsec:moe-analysis-appendix}

\begin{figure}[H]
\centering
  \includegraphics[width=\linewidth]{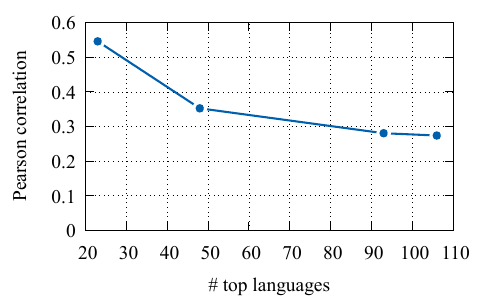}
  \caption{Pearson correlation values for distance between languages based on phylogenetic features and \approach's \moe routing features. The x-axis represents the numbers of languages included in the comparison. We include the languages in the descending order of the number of documents seen by the model for that language.}
  \label{fig:moe-gen-corr}
\end{figure}

\begin{figure*}[h]
\centering
  \includegraphics[width=\linewidth]{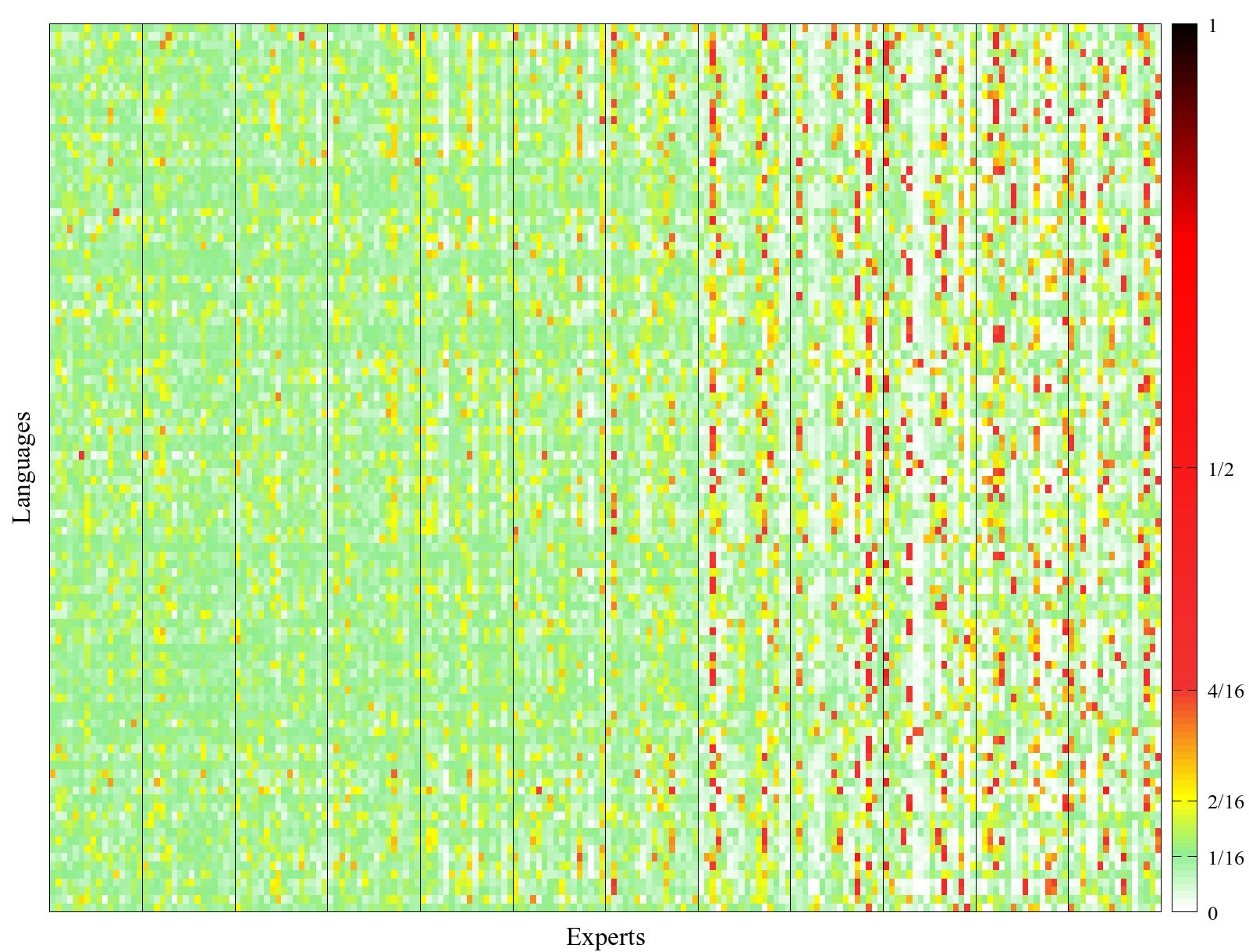}
  \caption{Heatmap showing the ratio of tokens routed to each expert across our model's layers. Each row represents a specific language, and each column corresponds to an expert. The heatmap tracks tokens from 106 languages as they pass through 12 \moe layers of the model, where an expert is assigned to each token in every layer. Vertical lines separate the 12 layers, ordered according to their position in the model, with 16 experts within each layer (from left to right).}
  \label{fig:moe-exp-heatmap}
\end{figure*}

The primary objective of this analysis is to explore the correlation between the expert vectors of \approach for each supported language and the corresponding language family groups. As discussed in \autoref{subsec:moe-analysis-main}, these vectors are derived from \approach's expert routing decisions for each language. To obtain them, we pass 10,000 sequences from each language through the model and record the number of tokens assigned to each expert. First, we normalize the vectors based on the norm of each layer, allowing us to determine whether certain experts exhibit specificity towards particular languages. As illustrated in \autoref{fig:moe-exp-heatmap}, the experts in the initial layers show less specificity, distributing tokens relatively evenly. However, in the later Transformer layers (closer to the output layer), token assignments seem to concentrate more heavily on certain experts. Upon closer inspection, we find that some of these experts display specificity for tokens from related languages. 

\begin{figure*}[h]
\centering
  \includegraphics[width=\linewidth]{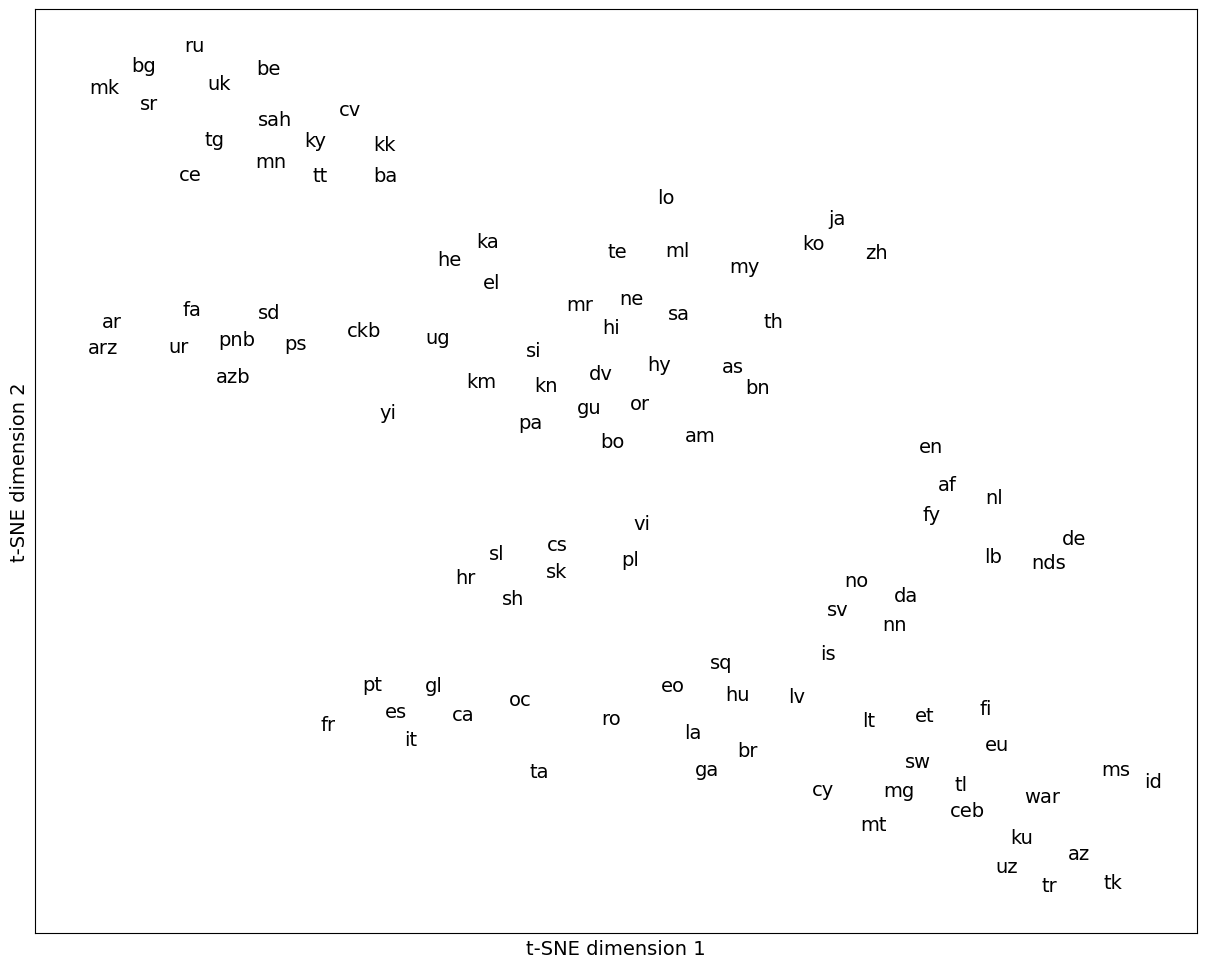}
  \caption{Two-dimensional t-SNE representation of the (normalized) expert-routing vectors obtained for each language through \approach. The language codes used are from \href{https://huggingface.co/datasets/uonlp/CulturaX\#languages}{CulturaX} dataset.}
  \label{fig:moe-tsne}
\end{figure*}

To investigate this phenomenon further, we use t-SNE representation after normalizing the vectors across all dimensions. As shown in \autoref{fig:moe-tsne}, many languages that share common language families are indeed clustered together. For instance: \begin{enumerate*} \item Romance languages such as French, Portuguese, Spanish, Italian, Galician, and Catalan; \item Indo-Aryan languages such as Hindi, Nepali, Marathi, and Sanskrit; \item Slavic languages such as Russian, Ukrainian, Belarusian, Bulgarian, Macedonian, and Serbian; and \item Germanic languages such as English, Afrikaans, Western Frisian, Dutch, German, Low German, and Luxembourgish. \end{enumerate*} However, we also identify some outliers, or false positives, such as: \begin{enumerate*} \item Korean, Japanese, and Chinese; and \item Vietnamese and Polish. \end{enumerate*}

This noise may stem from the t-SNE method itself. Thus, to validate these findings more formally, we further investigate the correlation between language family distances and the distances obtained from our expert-routing vectors, as outlined in \autoref{subsec:moe-analysis-main}. \autoref{fig:moe-gen-corr} demonstrates how the correlation values change when filtering the number of languages based on their proportion in the training data. The four data points in the figure are for the 23, 48, 93 and 106 languages that have at least 1000000, 100000, 10000, or 1000 documents in the training data, respectively.

\clearpage
\clearpage
\phantomsection
\label{sec:langdoc}

\tablehead{
  \toprule
  \textbf{Code} & \textbf{Language} & \textbf{Doc. Count} \\
  \midrule
}
\tabletail{
  \midrule
  \multicolumn{3}{r}{\small\slshape Continued} \\
  \bottomrule
}
\tablelasttail{
}
\begin{supertabular}{p{0.2\columnwidth}p{0.4\columnwidth}S[table-format=9]}
af   & Afrikaans         & 33060            \\
als  & Swiss German      & 6936             \\
am   & Amharic           & 9733             \\
an   & Aragonese         & 2746             \\
ar   & Arabic            & 2961118          \\
arz  & Egyptian Arabic   & 71625            \\
as   & Assamese          & 52627            \\
ast  & Asturian          & 9002             \\
av   & Avaric            & 438              \\
az   & Azerbaijani       & 203380           \\
azb  & South Azerbaijani & 29833            \\
ba   & Bashkir           & 71957            \\
bar  & Bavarian          & 3                \\
bcl  & Central Bikol     & 1                \\
be   & Belarusian        & 65739            \\
bg   & Bulgarian         & 965272           \\
bh   & Bihari languages  & 265              \\
bn   & Bangla            & 497463           \\
bo   & Tibetan           & 54185            \\
bpy  & Bishnupriya       & 5087             \\
br   & Breton            & 43765            \\
bs   & Bosnian           & 1237             \\
bxr  & Russia Buriat     & 100              \\
ca   & Catalan           & 621271           \\
cbk  & Chavacano         & 2                \\
ce   & Chechen           & 17322            \\
ceb  & Cebuano           & 10555            \\
ckb  & Central Kurdish   & 6881             \\
cs   & Czech             & 2614022          \\
cv   & Chuvash           & 22570            \\
cy   & Welsh             & 21998            \\
da   & Danish            & 1017192          \\
de   & German            & 67202797         \\
dsb  & Lower Sorbian     & 59               \\
dv   & Divehi            & 66702            \\
el   & Greek             & 2057209          \\
eml  & Emiliano-Romagnol & 91               \\
en   & English           & 64821313         \\
eo   & Esperanto         & 18403            \\
es   & Spanish           & 18037505         \\
et   & Estonian          & 320190           \\
eu   & Basque            & 63952            \\
fa   & Persian           & 2381245          \\
fi   & Finnish           & 1218706          \\
fr   & French            & 14550173         \\
frr  & Northern Frisian  & 11               \\
fy   & Western Frisian   & 8930             \\
ga   & Irish             & 12170            \\
gd   & Scottish Gaelic   & 8408             \\
gl   & Galician          & 71438            \\
gn   & Guarani           & 103              \\
gom  & Goan Konkani      & 721              \\
gu   & Gujarati          & 46515            \\
he   & Hebrew            & 186159           \\
hi   & Hindi             & 786614           \\
hr   & Croatian          & 18427            \\
hsb  & Upper Sorbian     & 4244             \\
ht   & Haitian Creole    & 12               \\
hu   & Hungarian         & 1765286          \\
hy   & Armenian          & 118579           \\
ia   & Interlingua       & 613              \\
id   & Indonesian        & 930054           \\
ie   & Interlingue       & 4                \\
ilo  & Iloko             & 2328             \\
io   & Ido               & 1144             \\
is   & Icelandic         & 94942            \\
it   & Italian           & 8452396          \\
ja   & Japanese          & 4447539          \\
jbo  & Lojban            & 1349             \\
jv   & Javanese          & 2058             \\
ka   & Georgian          & 124812           \\
kk   & Kazakh            & 109359           \\
km   & Khmer             & 40527            \\
kn   & Kannada           & 54085            \\
ko   & Korean            & 822292           \\
krc  & Karachay-Balkar   & 1745             \\
ku   & Kurdish           & 11812            \\
kv   & Komi              & 1396             \\
kw   & Cornish           & 94               \\
ky   & Kyrgyz            & 22836            \\
la   & Latin             & 48968            \\
lb   & Luxembourgish     & 6635             \\
lez  & Lezghian          & 1806             \\
li   & Limburgish        & 206              \\
lmo  & Lombard           & 3530             \\
lo   & Lao               & 8713             \\
lrc  & Northern Luri     & 43               \\
lt   & Lithuanian        & 533591           \\
lv   & Latvian           & 285463           \\
mai  & Maithili          & 93               \\
mg   & Malagasy          & 4636             \\
mhr  & Eastern Mari      & 7883             \\
min  & Minangkabau       & 1429             \\
mk   & Macedonian        & 110512           \\
ml   & Malayalam         & 107722           \\
mn   & Mongolian         & 77153            \\
mr   & Marathi           & 90663            \\
mrj  & Western Mari      & 1056             \\
ms   & Malay             & 9526             \\
mt   & Maltese           & 6052             \\
mwl  & Mirandese         & 9                \\
my   & Burmese           & 34623            \\
myv  & Erzya             & 4                \\
mzn  & Mazanderani       & 1914             \\
nah  & Nahuatl languages & 131              \\
nap  & Neapolitan        & 31               \\
nds  & Low German        & 15139            \\
ne   & Nepali            & 124961           \\
new  & Newari            & 4344             \\
nl   & Dutch             & 4695706          \\
nn   & Norwegian Nynorsk & 5043             \\
no   & Norwegian         & 756292           \\
oc   & Occitan           & 10556            \\
or   & Odia              & 6138             \\
os   & Ossetic           & 8596             \\
pa   & Punjabi           & 25879            \\
pam  & Pampanga          & 4                \\
pl   & Polish            & 5686688          \\
pms  & Piedmontese       & 7566             \\
pnb  & Western Panjabi   & 15625            \\
ps   & Pashto            & 15076            \\
pt   & Portuguese        & 7611586          \\
qu   & Quechua           & 1202             \\
rm   & Romansh           & 30               \\
ro   & Romanian          & 1613016          \\
ru   & Russian           & 31972436         \\
rue  & Rusyn             & 1                \\
sa   & Sanskrit          & 16290            \\
sah  & Sakha             & 22141            \\
scn  & Sicilian          & 21               \\
sd   & Sindhi            & 4366             \\
sh   & Serbian (Latin)   & 45619            \\
si   & Sinhala           & 30146            \\
sk   & Slovak            & 743300           \\
sl   & Slovenian         & 293415           \\
so   & Somali            & 39               \\
sq   & Albanian          & 208223           \\
sr   & Serbian           & 162126           \\
su   & Sundanese         & 1554             \\
sv   & Swedish           & 1988367          \\
sw   & Swahili           & 66506            \\
ta   & Tamil             & 189138           \\
te   & Telugu            & 72914            \\
tg   & Tajik             & 19353            \\
th   & Thai              & 838422           \\
tk   & Turkmen           & 14393            \\
tl   & Filipino          & 13938            \\
tr   & Turkish           & 3768298          \\
tt   & Tatar             & 8724             \\
tyv  & Tuvinian          & 23               \\
ug   & Uyghur            & 47035            \\
uk   & Ukrainian         & 1789621          \\
ur   & Urdu              & 110291           \\
uz   & Uzbek             & 87219            \\
vec  & Venetian          & 113              \\
vi   & Vietnamese        & 4096447          \\
vls  & West Flemish      & 1                \\
vo   & Volapük           & 6621             \\
wa   & Walloon           & 1383             \\
war  & Waray             & 23687            \\
wuu  & Wu Chinese        & 222              \\
xal  & Kalmyk            & 51               \\
xmf  & Mingrelian        & 9706             \\
yi   & Yiddish           & 5646             \\
yo   & Yoruba            & 192              \\
yue  & Yue Chinese       & 3                \\
zh   & Chinese           & 8744984          \\
\midrule
\multicolumn{2}{l}{Total Doc. Count} & 275653546 \\
\bottomrule
\end{supertabular}
\vspace{0.5em}
\captionof{table}{List of languages included in the CulturaX dataset, along with the corresponding number of documents per language in the training sample used for \approach. The language codes utilized are derived from the CulturaX dataset, which adheres to a combination of ISO 639-1 and ISO 639-3 standards. An exception is the use of \textit{als}, which is considered obsolete; the ISO 639-3 standard designates \textit{gsw} as its replacement.}



\end{document}